\documentclass{article}

\usepackage{arxiv}

\usepackage[utf8]{inputenc} 
\usepackage[T1]{fontenc}    
\usepackage{hyperref}       
\usepackage{url}            
\usepackage{booktabs}       
\usepackage{multirow}
\usepackage{amsfonts}       
\usepackage{nicefrac}       
\usepackage{microtype}      
\usepackage{lipsum}		
\usepackage{graphicx}
\usepackage{natbib}
\usepackage{doi}
\usepackage{amssymb}

\usepackage{xcolor}
\usepackage{ifthen}
\usepackage{multicol}
\usepackage{tcolorbox}
\usepackage{amsmath}
\usepackage[misc]{ifsym}

\newcommand{\submitmode}{false} 
\ifthenelse{\equal{\submitmode}{true}}{
	\newcommand{\rmk}[1]{}
	\newcommand{\rmkdone}[1]{}
	\newcommand{\ldel}[1]{}
	
	\newcommand{\note}[1]{}
}
{
	\newcommand{\rmk}[1]{\textcolor{orange}{--[\textbf{#1}}]}
	\newcommand{\rmkdone}[1]{}
	
	\newcommand{\ldel}[1]{\textcolor{cyan}{\sout{#1}}}
	\newcommand{\note}[1]{\textcolor{green}{#1}}
}

\title{Eden: A Unified Environment Framework for Booming Reinforcement Learning Algorithms}

\author{Ruizhi Chen$^1$, Xiaoyu Wu$^1$, Yansong Pan$^2$, Kaizhao Yuan$^2$, Ling Li$^1$, TianYun Ma$^3$, JiYuan Liang$^3$, 
	\\\textbf{Rui Zhang}$^2$,\textbf{Kai Wang}$^2$, \textbf{Chen Zhang}$^3$, \textbf{Shaohui Peng}$^2$, \textbf{Xishan Zhang}$^2$, \textbf{Zidong Du}$^2$, \textbf{Qi Guo}$^2$,
	\\ \Letter\space \textbf{Yunji Chen}$^{2,3}$
	\\ 1 Institute of Software Chinese Academy of Sciences
	\\ 2 Institute of Comuting and Technology Chinese Academy of Sciences
	\\ 3 University of Science and Technology of China
	\\ \Letter \space cyj@ict.ac.cn}

\renewcommand{\shorttitle}{\textit{arXiv} Template}

\hypersetup{
	pdftitle={A template for the arxiv style},
	pdfsubject={q-bio.NC, q-bio.QM},
	pdfauthor={David S.~Hippocampus, Elias D.~Striatum},
	pdfkeywords={First keyword, Second keyword, More},
}

\begin{document}
\maketitle

\begin{abstract}
	With AlphaGo defeats top human players, reinforcement learning(RL) algorithms have gradually become the code-base of building stronger artificial intelligence(AI). The RL algorithm design firstly needs to adapt to the specific environment, so the designed environment guides the rapid and profound development of RL algorithms. However, the existing environments, which can be divided into real world games and customized toy environments, have obvious shortcomings. For real world games, it is designed for human entertainment, and too much difficult for most of RL researchers. For customized toy environments, there is no widely accepted unified evaluation standard for all RL algorithms. Therefore, we introduce the first virtual user-friendly environment framework for RL. In this framework, the environment can be easily configured to realize all kinds of RL tasks in the mainstream research. Then all the mainstream state-of-the-art(SOTA) RL algorithms can be conveniently evaluated and compared. Therefore, our contributions mainly includes the following aspects: 1.single configured environment for all classification of SOTA RL algorithms; 2.combined environment of more than one classification RL algorithms; 3.the evaluation standard for all kinds of RL algorithms. With all these efforts, a possibility for breeding an AI with capability of general competency in a variety of tasks is provided, and maybe it will open up a new chapter for AI. 
\end{abstract}


\section{Introduction}

\begin{figure*}[htbp]
	\centering
	\includegraphics[width=0.8\textwidth]{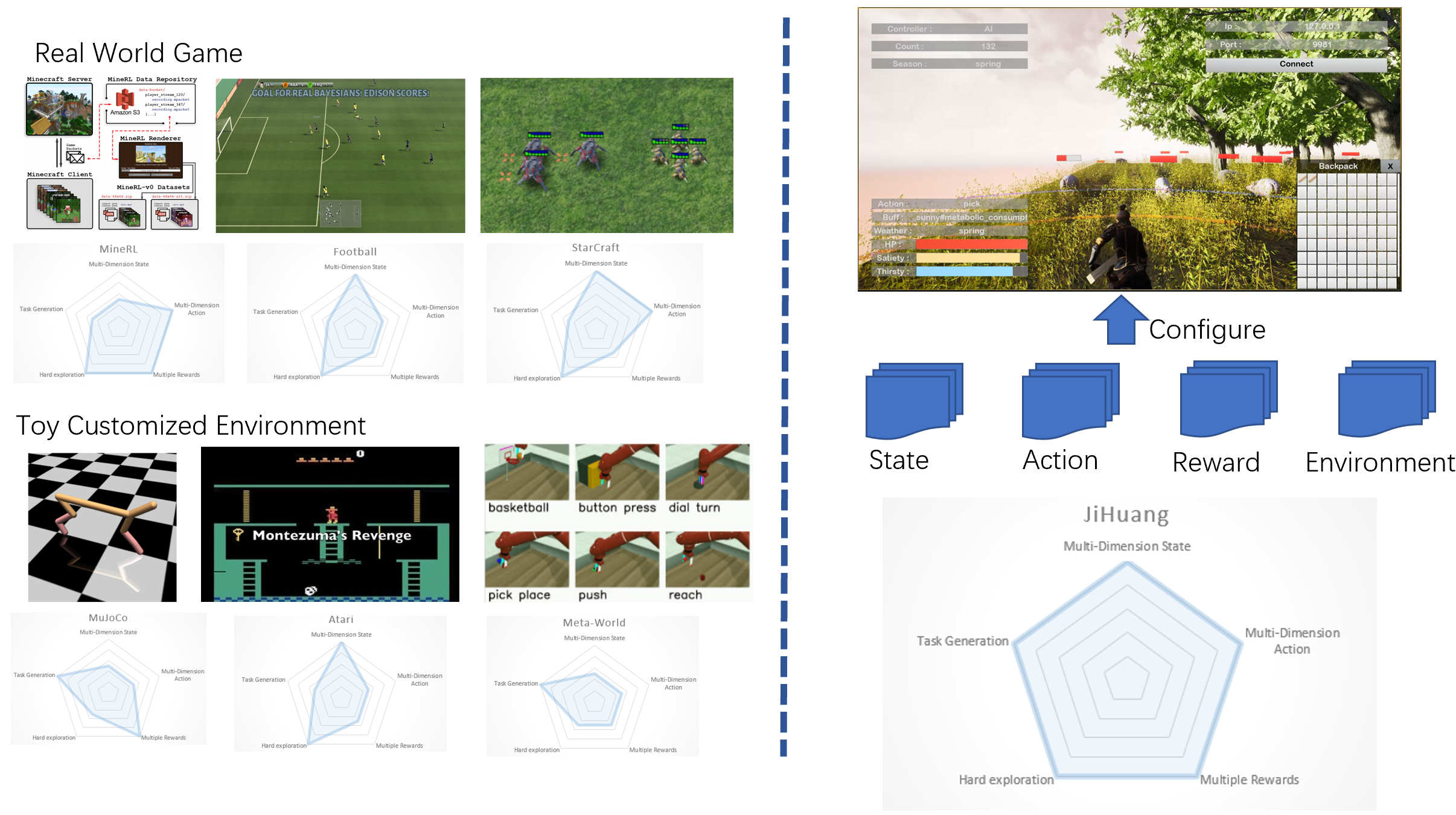}
	\caption{The configurability of different environments.}
	\label{fig:figure1}
\end{figure*}

In recent years, deep reinforcement learning (DRL) algorithms have achieved significant progress based on real-world games or customized toy environments. The real-world games are computer games designed for human entertainment in the real world, such as AlphaGo\cite{SilverSchrittwieser-67} and OpenAI Five\cite{berner2019dota}. Such environments are too complex, and require huge computing resources for DRL development. The customized toy environments are simple environments customized for some specific problems, such as Meta-World\cite{YuQuillen-8} and CityFlow\cite{ZhangFeng-86}. Such environments can only be used to test some specific reinforcement learning with limited action or reward or state spaces, as shown in Figure~\ref{fig:figure1}. Therefore, existing environments are not suitable for developing reinforcement learning algorithms with limited computing resources for various problems towards general artificial intelligence. And they are not suitable for evaluating different reinforcement learning algorithms either.

To provide a general evaluation platform, some configurable environments have been proposed, such as XLand\cite{team2021open} and Arena\cite{SongWojcicki-85}. XLand is designed to evolve an agent with better generalization performance. However, XLand could not be used to evaluate various reinforcement learning algorithms designed for various fields, such as Meta-RL\cite{YuQuillen-8}, Batch RL\cite{fujimoto2019benchmarking}, etc. Arena is a general platform designed for multi-agent reinforcement learning. It emphasizes the measurement of competition and cooperation between agents, and at the same time constructs a Social Tree that expresses this measurement. However, Arena is not suitable for other kinds of reinforcement learning, such as Exploration Strategies\cite{mcfarlane2018survey}, Model-Based RL\cite{moerland2020model}.

To evaluate and develop DRL algorithms conveniently, we propose the first virtual RL-friendly environment framework (Eden). We use C++ to build the back-end logic of the game, and support the use of a simple configuration file to define the distribution of various resources in the instantiation environment. At the same time, we provide Gym interface and corresponding Wrapper to configure the environment to meet different algorithm requirements.
With Eden, environments with different reward or state or action spaces could be configured to boom various DRL algorithms. 
First, all representative RL algorithms should be easily trained and evaluated on our environment. So our environment should be configured as a single environment that only contains RL tasks that logically equal to each representative algorithms. To ensure the equivalence, similar computational results should be evaluated on these configured environments. Moreover, our environment can be more difficult than the environment used in the representative algorithms through adjusting the environment parameters. Second, the configured environments mentioned above can be combined into a more complex environment so that more RL techniques should be introduced in one RL algorithm to extend the ability of RL algorithms. This algorithm with more RL techniques can be seen as the stronger AI than the representative algorithms. Meanwhile, the most complex and general environment in our study is the environment that contains all the specific tasks of all representative RL algorithms. Third, all kinds of RL algorithms in different RL classification can be evaluated in one configured environment of our framework. If the single configured environment contains one specific RL task and be used to evaluate one representative algorithm, then the combined configured environment can be used to evaluate the other ability of the representative algorithm.

\section{Background and Motivation}
In this section, we first introduced the basic framework of RL, then introduced the difficulties of building an AI system in the real-world games, and finally introduced the relevant characteristics of the RL algorithm used in the customized toy environments.

\subsection{Foundation of RL}
\begin{figure}[htbp]
	\centering
	\includegraphics[width=0.5\textwidth]{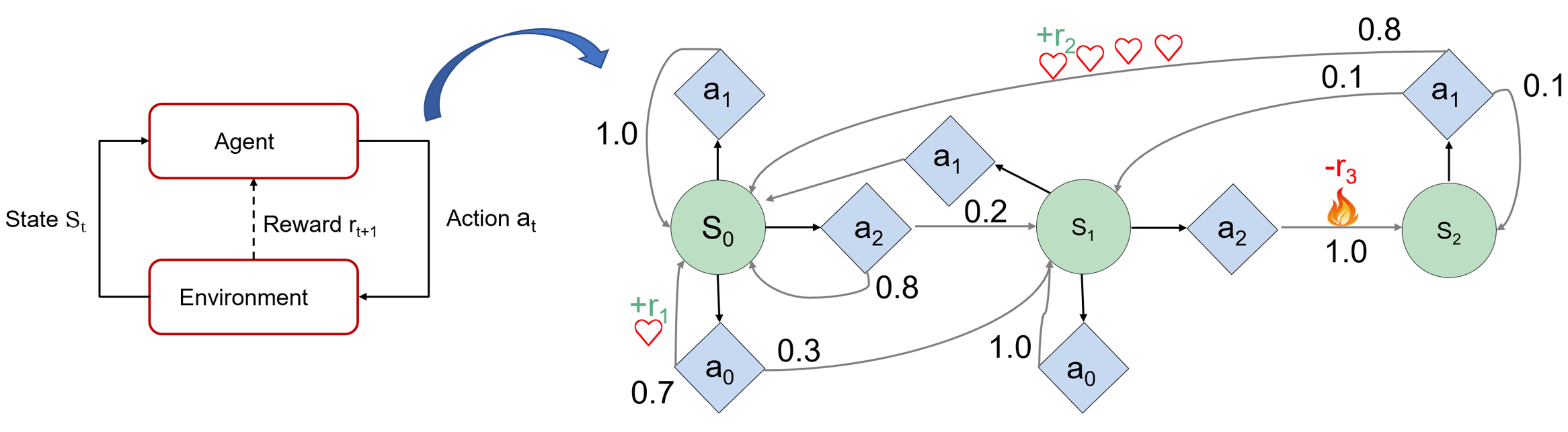}
	\caption{The standard framework of RL.}
	\label{fig:mdp}
\end{figure}

Reinforcement learning is that the agent interacts with the environment and learns what to do, how to map states to actions, so as to maximize the expected reward\cite{SuttonBarto-39}. The standard framework of RL is shown in the left side of Figure~\ref{fig:mdp}. At time step $t$, the agent observes a state $s_t\in \mathcal{S}$ from the environment and interacts an action $a_t\in \mathcal{A}$, where $\mathcal{S}$ is the state space and $\mathcal{A}$ is the action space. Then, the environment will push forward one time step with a transition  function $\mathcal{T}: \mathcal{S}\times\mathcal{A}\rightarrow\mathcal{S}$ and output a reward $r_{t+1}$ with a function $\mathcal{R}:\mathcal{S}\times\mathcal{A}\times\mathcal{S}\rightarrow\mathbb{R}$. In this framework, the agent is usually modeled with the Markov Decision Process(MDP), as shown in the right side of Figure~\ref{fig:mdp}. The goal of the agent is to obtain the maximum expected cumulative funture return $G_t = E_{\pi}[\sum_{t=0}^{\infty}\gamma^t\mathcal{R}(s_t,a_t,s_{t+1})]$ through the trained policy $\pi(a_t|s_t)$. From the example of Figure~\ref{fig:mdp}, suppose there are two different polices $\pi_1$ and $\pi_2$($\gamma=0.9$), policy $\pi_1$ satisfies that $\pi_1(a_0|s_0)=1$ and $\pi_1(a_1|s_1)=1$, and policy $\pi_2$ satisfies that $\pi_2(a_2|s_0)=1$, $\pi_2(a_2|s_1)=1$ and $\pi_2(a_1|s_2)=1$. Based on the above assumptions, the expected return of policy $\pi_1$ is $\frac{0.7r_1}{1-0.7\gamma-0.3\gamma^2}=5.51r_1$, the expected return of policy $\pi_2$ is $\frac{8\gamma^2r_2-(10\gamma-\gamma^2)r_3}{(1-\gamma)(50+5\gamma+4\gamma^2)}=1.12r_2-1.42r_3$. Therefore, the value of $r_1,r_2,r_3$ decide whether $\pi_1$ is better than $\pi_2$. Moreover, it is much harder to achieve $\pi_2$ because of the longer actoin sequence. In summary, the policy of the agent is highly relative with the design of the environment.  

\subsection{Real World Game}

Real world games is the computer games designed for human entertainment, and many researches have focuses on these computer games. To a certain extend, these games have certain similarities with some parts of the real world. For example, StarCraft simulates the war between different races for limited sources\cite{vinyals2019grandmaster}. Meanwhile, the existing RL algorithm framework requires a large amount of data for trial-and-error training\cite{SuttonBarto-39}, but the acquisition of real world data has the disadvantages of high acquisition cost and limited data quantity. On the other hand, real world games can cheaply simulate an infinite amount of training data while showing some parts of characteristics of the real world. Therefore, many scientific research institutions are committed to developing AI with super-human performance in real world games, such as AlphaGo\cite{SilverSchrittwieser-67}, AlphaStar\cite{vinyals2019grandmaster}, FTW\cite{jaderberg2019human} and so on.

For small and medium-sized scientific research institutions, it is very difficult to build an AI system that can defeat top human players in real-world games. Take AlphaStar as an example\cite{vinyals2019grandmaster}, the difficulties of the StarCraft game itself include the large number of action space combinations, incomplete information, and only one sparse reward for tens of thousands of game steps. For the built AI system, if you simply use the self-play method to learn strategies, there may be strategy cycles during the training process. Furthermore, the learned strategies may not be effective against human strategies. Therefore, the techniques of self-attention, scatter connection, LSTM, auto-regressive policy and recurrent pointer network are introduced to contruct the neural networks in this AI system. During the training process, supervised learning techniques are used to train and predict the next action through human data and reinforcement learning algorithms are used to maximize the winning rate. For the reinforcement learning algorithms, the techniques of A3C, V-trace, TD($\lambda$), UPGO are used to build the RL framework. To prevent strategy cycles, the league training is introduced to choose the suitable opponents for improving AI performance. Above all the techniques used, AlphaStar is a complex and sophisticated AI system with various engineering tricks.

Although some sophisticated AI systems have built on some real world games, more reinforcement learning techniques should be added to these AI systems. Only few RL techniques have been chosen in these AI systems, and most of the RL algorithms do not be chosen. There may be some suitable algorithms, but it is not the right time to add them to these AI systems now. With the development of RL algorithms, some parts of these systems can obviously be replaced with some better-performing technologies. However, none of these systems have been open sourced, and then it is hard to construct better AI systems based on the above AI systems. Meanwhile, the customized toy environments used in the algorithm papers are too simple, and there is no need to build the above AI systems on these environments. A natural idea is to build a configurable environment. Through simple configuration, this environment can run existing algorithms. When more complex AI systems need to be studied, this environment can be expanded to approach the difficulty of real-world games. The configurability of some games is shown in Table~\ref{tab:configurability}.

\begin{table*}[!htbp]
	\centering
	\caption{The configurability of different RL environments}
	\label{tab:configurability}
	\begin{tabular}{|l|c|c|c|c|c|c|c|c|}
		\hline
		&  Atari & MuJoCo & Meta-World & MineRL & Football & StarCraft & RLWorld \\ \hline
		Multi-Dimension State &  $\surd$ & $\times$ & $\times$ & $\times$ & $\surd$ & $\surd$ & $\surd$ \\ \hline
		Multi-Dimension Action &  $\times$ & $\times$ & $\times$ & $\surd$ & $\times$ & $\surd$ & $\surd$ \\ \hline
		Multiple Rewards &  $\times$  & $\surd$ & $\times$ & $\surd$ & $\times$ & $\times$ & $\surd$ \\ \hline
		Hard Exploration &  $\surd$ & $\times$ & $\times$ & $\surd$ & $\surd$ & $\surd$ & $\surd$ \\ \hline
		Task Distribution  & $\times$ & $\times$ & $\surd$ & $\times$ & $\times$ & $\times$ & $\surd$ \\ \hline
	\end{tabular}
\end{table*}

\subsection{Customized Toy Environment}
Customized toy environment is the most widely used environment in RL. Each environment is designed to show the superiority of the proposed algorithm in each research. Therefore, all these environments should contain as many suitable difficult problems as it could be for the relative capacity of the proposed algorithm. In this paper, we call these similar difficult problems in one kind of customized toy environment as one functional characteristic of the environment. In order to summarize these functional characteristics, we use RL algorithm classification to determine the difficult problems solved by each kind of RL algorithms. The result is shown in Table~\ref{table:toy_environment}.

\begin{table*}[!htbp]
	\centering
	\caption{Different category of reinforcement learning algorithms and their test environments.}
	\label{table:toy_environment}
	\begin{tabular}{|l|c|l|l|}
		\hline
		& Current Area & Contribution & Environment \\ \hline
		\multirow{4}{*}{Problem} & Model-Free & Basic solution & \begin{tabular}[c]{@{}l@{}}1.MuJoCo locomotion\\ 2.Atari\\ 3.GridWorld\end{tabular} \\ \cline{2-4} 
		& Exploration & \begin{tabular}[c]{@{}l@{}}1.Sparse reward\\ 2.Hard Exploration\end{tabular} & \begin{tabular}[c]{@{}l@{}}1.Maze/Navigation\\ 2.Atari(Montezuma)\end{tabular} \\ \cline{2-4} 
		& Multitask RL & Learn one policy for multiple tasks & 1.Meta-World \\ \cline{2-4} 
		& Multi-Agent & \begin{tabular}[c]{@{}l@{}}1.Multi-agent credit assignment\\ 2.Non-stationarity\\ 3.Curse of dimensionality\\ 4.Paritial observability\end{tabular} & \begin{tabular}[c]{@{}l@{}}1.StarCraft II\\ 2.Quake III\\ 3.Google Football\\ 4.CityFlow\\ 5.NeuralMMO\end{tabular} \\ \hline
		\multirow{5}{*}{Technique} & Meta-RL & Leverage prior knowledge to new tasks & \begin{tabular}[c]{@{}l@{}}1.Robotic Manipulation\\ 2.Navigation/Locomotion\\ 3.Meta-World\end{tabular} \\ \cline{2-4} 
		& Hierarchical RL & Multiple complicated tasks & 1.Ant Maze \\ \cline{2-4} 
		& Model-Based & \begin{tabular}[c]{@{}l@{}}1.Dynamics bottleneck\\ 2.Planning horizon dilemma\\ 3.Early termination dilemma\end{tabular} & \begin{tabular}[c]{@{}l@{}}1.MuJoCo\\ 2.DeepMind Control Suite\end{tabular} \\ \cline{2-4} 
		& Batch RL & Extrapolation error & Atari \\ \cline{2-4} 
		& Multiobjective RL & Pareto front & \begin{tabular}[c]{@{}l@{}}1.Resource Gathering\\ 2.MO-Mountain-Car etc\end{tabular} \\ \hline
	\end{tabular}
\end{table*}

\subsubsection{Model-Free RL}
Model-Free RL is one of the most widely used algorithms. These kinds of algorithms can be divided into two main approaches: Policy Optimization and Q-Learning. Polciy Optimization methods learn an actor as an optimal policy $\pi^{*}(a_t|s_t)$, which satisfies that
\begin{equation}
\pi^{*} = argmax_{\pi_{\theta}}\mathbb{E}_{(s_t,a_t)\sim\pi_{\theta}}[\sum_tR(s_t,a_t)]
\end{equation}
These methods directly optimize the objective for the thing the agent wants, which tends to make them reliable and stable. The typical algorithms are Policy Gradient\cite{SuttonBarto-39}, A2C/ A3C\cite{MnihBadia-14}, TRPO\cite{SchulmanLevine-34}, PPO\cite{SchulmanWolski-15} etc. Q-Learning methods learn a critic that approximates the optimal action-value function 
\begin{equation}
Q^*(s_t,a_t)=max_{\pi}\mathbb{E}[R(s_t,a_t)+\gamma R(s_{t+1},a_{t+1})+...|s_t=s,a_t=a,\pi] 
\end{equation}
These methods indirectly optimize the agent objective throgh a self-consistency action-value equation, whcih tends to be less stable but substantially more sample efficient if they do work. The typical algorithms of are DQN\cite{MnihKavukcuoglu-16,MnihKavukcuoglu-17}, C51\cite{BellemareDabney-37}, QR-DQN\cite{DabneyRowland-38}, HER\cite{AndrychowiczWolski-18} etc. Therefore, some works interpolate between Policy Optimization and Q-Learning, such as DDPG\cite{LillicrapHunt-35}, TD3\cite{FujimotoHoof-36}, SAC\cite{HaarnojaZhou-19} etc.

\subsubsection{Exploration Strategies}
Exploration versus exploitation is an essential and fundamental dilemma through the development of RL algorithms. On the one hand, the RL agent should exploit the trained policy to choose optimal actions to achieve more environmental rewards as fast as possible. However, on the other hand, the agent should try out some nonoptimal actions to explore the environment reserving the possiblity of obtaining more rewards latter from the unexplored environment states. Without enough exploration, the agent should lead to local minima or total failure. Unfortunately, the tradeoff between exploration and exploitation is strongly coupled with the specific environment. For example, the agent could make great results in most Atari games without much exploration. In the meantime, some Atari games, such as Montezuma's Revenge and Pitfull, need much more exploration to get positive rewards. In a word, how to achieve good exploration quite efficiently without the specific environment will always be an open topic.

The exploration strategies can be divided into four main categories. First, some classic exploration strategies works well through the RL development. In simple tabular RL, the techniques, such as epsilon-greedy\cite{MnihKavukcuoglu-16,MnihKavukcuoglu-17}, upper confidence bounds, work out pretty well. When neural networks fit the job of function approximation, the following techniques, such as entropy loss term\cite{HaarnojaZhou-19} and noise-based exploration\cite{FortunatoAzar-47}, are chosen for better exploration. Second, intrinsic reward, which is inspired by intrinsic motivation in psychology, is one common approach to augment the environment reward for better extra exploration. Then, the policy is trained by the composed reward as
\begin{equation}
r_t = r_t^e+\beta r_t^i
\end{equation}
where $r_t^e$ denotes the environment reward and $r_t^i$ denotes the intrinsic reward. These approaches can be split into two groups, count-based exploartion\cite{BellemareSrinivasan-40,OstrovskiBellemare-41,TangHouthooft-42} and prediction-based model\cite{OudeyerKaplan-43,PathakAgrawal-21,BurdaEdwards-22,PathakGandhi-49}. The aim of count-based exploration is to discover novel states, and it can be grouped through the count methods, such as counting by density model\cite{BellemareSrinivasan-40} and counting after Hashing\cite{TangHouthooft-42}. Prediction-based model adds the bounus intrinsic reward through the improvement of the agent environment knowledge. ICM\cite{PathakAgrawal-21} and RND\cite{BurdaEdwards-22} are influential works in this area. Another category of exploration strategies is memory-based exploration\cite{BadiaSprechmann-50,BadiaPiot-2,EcoffetHuizinga-7}. These methods often encourage exploration through external memory. In NGU\cite{BadiaSprechmann-50}, episodic memory is introduced to remember the knowledge between different agent episodes. In Go-Explore\cite{EcoffetHuizinga-7}, the running pathes are stored in an archive, and the agent can restart one path in the archive to extra exploration.

\subsubsection{Multitask RL}
Multitask RL refers to training on a group of tasks separately and then testing on arbitrary tasks the agent met in training. The core objective is to leverage the domain-related information accumulated by training the individual, related tasks in parallel with a shared representation of the system\cite{MaurerPontil-53}. The key of multitask RL is the performance bottleneck problems experienced by the agent are drawn from the same distribution. There are several challenges in this ar, such as distraction dilemma, catastrophic forgetting, and negative knowledge transfer\cite{VithayathilVargheseMahmoud-54}.

Various approaches have been developed for multitask RL, such as transfer learning-oriented approach, shared representations for value functions, progressive neural networks\cite{RusuRabinowitz-56}, PathNet\cite{FernandoBanarse-55}, policy distillation\cite{RusuColmenarejo-57}, and A3C\cite{VithayathilVargheseMahmoud-54}. Drawing support of RNN structures such as LSTM, these methodologies take care of the main problems in multitask RL, for instance, catastrophic forgetting. Along with the approaches above, Deepmind has also come up with several mature solutions for multitask RL, named DISTRAL\cite{TehBapst-58}, PoPArt\cite{HesselSoyer-59}, and IMPALA\cite{EspeholtSoyer-60}. The solutions either use the A3C model or a centroid policy model as their operating model and leverage different technologies to overcome the dilemma in multitask RL.

\subsubsection{Hierarchical RL}
When the environments become more complex, the basic RL architecture, as shown in Figure~\ref{fig:mdp}(a), suffers from a variety of defects that hinder RL learning, such as sparse reward and hard exploration. Hierarchical RL aims at breaking down specfic parts of complicated tasks, so called temporal abstraction, to alleviate the learning complexity. Temporal abstraction usually means the extension of the set of available actions through temporal domain. Therefore, the agent of Hierarchical RL can execute not only elementory actions, but also macro-actions, which denote sequences of elementary actions. Indeed, Hierarchical RL often train a policy made up of multiple layers, each of which control at different level of temporal abstraction. The manager sets up the goal state at high level of temporal abstraction, and the worker tries to achieve this state with several elementary actions.

Hierarchical RL can usually be divided into four categories. First, the skilled-based methods, such as SSN4HRL\cite{FlorensaDuan-76}, MLSH\cite{FransHo-77}, defined shared action sequences at the lower layer and learn the control policy at the higher layers. Second, the multiple layer methods, such as HAC\cite{LevyKonidaris-78}, DEHRL\cite{SongWang-79}, study how to construct more than two abstract layers in Hierarchical RL. Third, the unsupervised hierarchical methods, such as HSP\cite{SukhbaatarDenton-80}, HAAR\cite{LiWang-81}, used some unsupervised methods to decide wether to add a new goal at the higher layers. Fourth, the methods would mix the above methods and other RL algorithms to achieve better performance at some simple environments.

\subsubsection{Meta-RL}
Though current deep RL has attained superb results in an expanding list of experiments taking advantage of the power of neural networks, it remains two main defeats: sample inefficient and narrow task adpation. Researchers, therefore, refer to approaches such as deep meta-RL to deal with the dilemma. Unlike regular RL, the objective of meta RL is to maximize the expected reward over a set of tasks, or Markov Decision Processes(MDPs), by learning the meta knowledge among them\cite{HuismanvanRijn-46}. The tasks are sampled from a same distribution $p(\mathcal{T})$, and then divided into train tasks set and test tasks set. The learning process encompass inner-level and outer-lever to finally find the task-specifical parameter $\phi$ via learning the meta parameter $\theta$.

Meta-RL chiefly contains two categories: model-based approaches and optimization-based approaches.Model-based approaches leverage outer or inner recurrent structures as memory. They perform well in small tasks set due to the flexiblity of the internal representaion. Several approaches such as SNAIL\cite{MishraRohaninejad-61} and RML\cite{WangKurth-Nelson-62} attain plausible result in simple RL environments. Optimization-based apporaches adopts directly fine tune instead. Regarding less of the task distributions, they achieves better results in a wider scope of RL problems than model-based approaches. MAML\cite{FinnAbbeel-28} is the most favorable one among them, and there are many impovement based on it, such as Reptile\cite{NicholAchiam-64}, BMAML\cite{YoonKim-65} and iMAML\cite{RajeswaranFinn-66}.

\subsubsection{Model-Based RL}
Model-Based RL is proposed to significantly improve sample efficiency of Model-Free RL. In Model-Based RL, the main problem can be divided into two categories: modeling and planning. In the modeling area, the algorithm concentrates on learning a proper transition dynamics of the environment as
\begin{equation}
\hat{s}_{t+1} \sim f_{\theta}(s_t,a_t)
\end{equation}
where $s_t$ and $a_t$ denote the current state and action respectively, $\hat{s}_{t+1}$ denotes the predicted state, and $f_{\theta}$ indicates the learned model. In the planning area, the basic assumption is that the environment model is available, and the algorithm concentrates on using the model to compute the next action as
\begin{equation}
a_{t+1} \sim \pi(s_t,f_{\theta})
\end{equation}
where $s_t$ denotes the current state, and $f_{\theta}$ indicates the learned model. Different from Model-Free RL, the policy $\pi$ is trained through maximizing the expected rewards over a predictive horizon of the environment model.

The Model-Based RL can be divided into three categories according to its usage. First, the model interacts with the agent to generate more data\cite{Sutton-68,KurutachClavera-70,KalweitBoedecker-71,GuLillicrap-72}, which can be used as an extra data source to accelerate the training process. Second, during Q-value or V-value estimation, the model is used as additional context information to help agent make decisions\cite{RacaniereWeber-74}. Third, the model is used to increase the quality of Q-value or V-value estimation in combination with Model-Free RL\cite{FeinbergWan-73,BuckmanHafner-75}.

\section{Properties of Eden}
In this section, we first introduce the basic properties of Eden, including the basic appearance of the world, the back-end subsystem of the environment framework, and two instantiation examples of Eden. Then we introduced the different configuration options of Eden and the experimental results of the corresponding configuration options. Finally, we introduced two kinds of evaluations to measure environmental difficulty.

\subsection{Foundation}
\begin{figure}[htbp]
	\centering
	\includegraphics[width=0.5\textwidth]{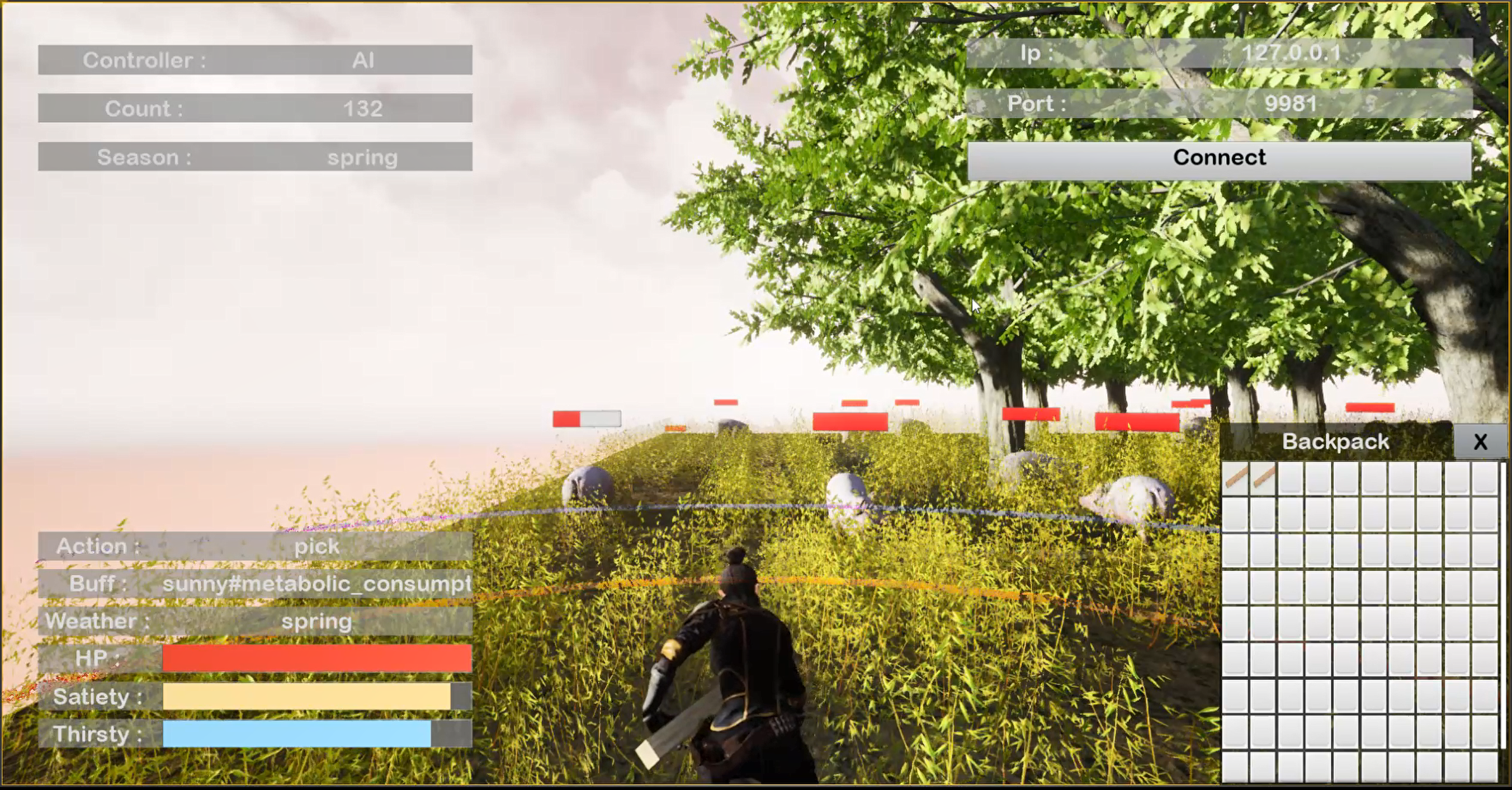}
	\caption{The display interface of the game. The upper left corner denotes the basic running status of the game, the upper right corner denotes the game server IP, and the lower left corner indicates the basic status information of the agent.}
	\label{fig:jihuang}
\end{figure}

In our environmental framework, the agent needs to survive in a configured wild world as long as possible. There are animals, plants and all kinds of survival tasks that Homo sapiens have encountered in real worlds, as shown in Figure~\ref{fig:jihuang}. The agent needs to learn how to capture different kinds of prey, collect useful materials from plants and animal carcasses, and use them to synthesize equipment needed in different seasons. While carrying out the above survival tasks, the agent should ensure their basic attributes are above a certain level. For example, the satiety of agent will decrease over time, and the agent needs to obtain animal meat through hunting to recover this basic attribute. The thirsty of agent, too, decrease over time, and agent needs to obtain water through collecting it from river. The upper limit of the satiety and thirsty are usually set to the same value, and is mentioned as life-limit. If the agent does nothing, it will live exactly for this life-limit timesteps.

\begin{figure}[htbp]
	\centering
	\includegraphics[width=0.5\textwidth]{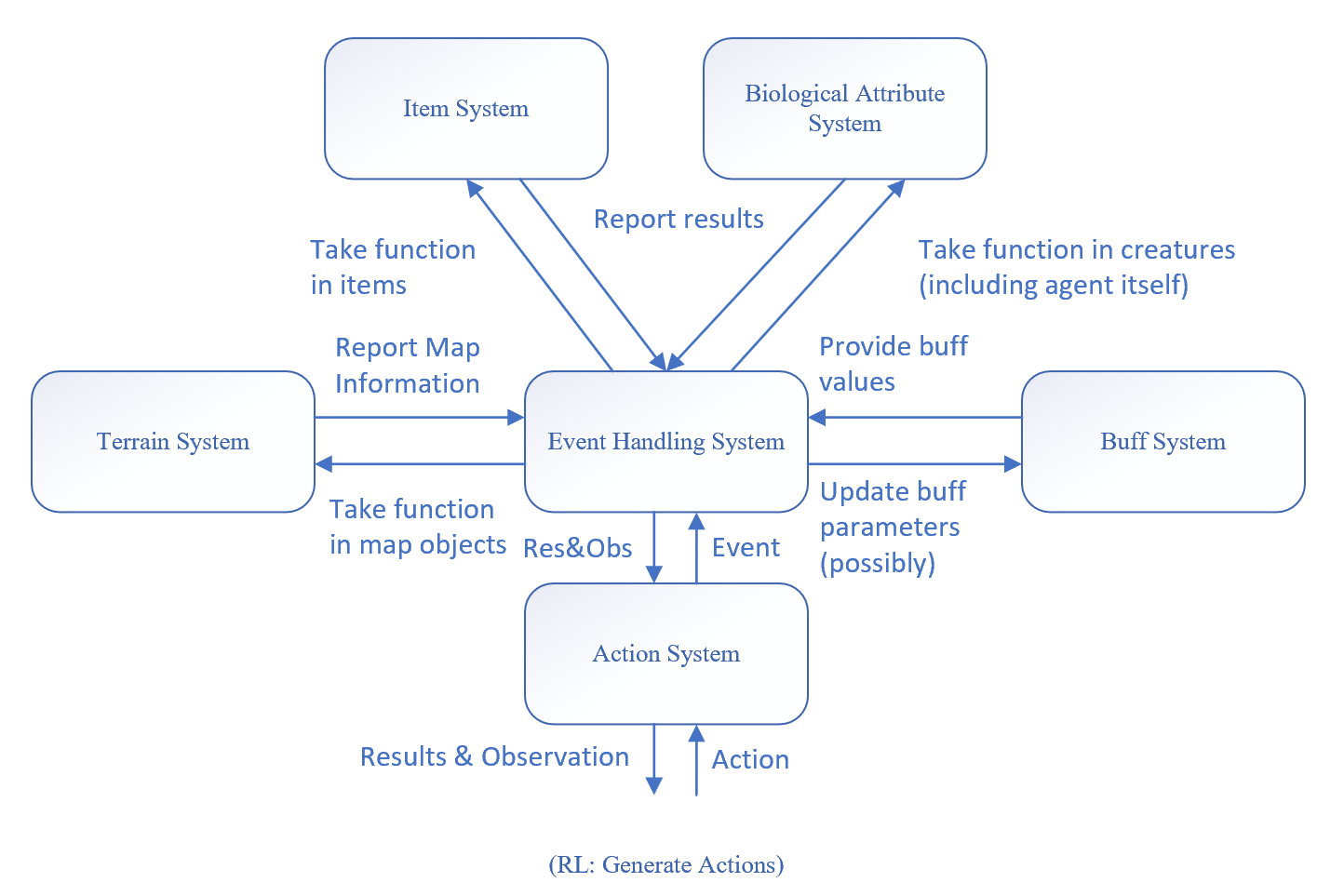}
	\caption{The six basic subsystems of the back-end game server.}
	\label{fig:six_system}
\end{figure}

Our environmental framework contains six main subsystems in the world game part, as shown in Figure~\ref{fig:six_system}. Action System defines the action used by the agent. It directly communicates with DRL, taking the algorithm output and returning the result of action and new observations. After getting the action, Action System generates an event to handle and hand it to Event Handling System. Biological Attribute System decides how the basic attributes of all the creatures change in our world. New attributes can be easily extended to our world through modifying the configuration file. Item System controls the change of equipment, materials, consumable items and so on. The synthesis table and the drop table in the configuration file can be modified to add new items in Item system. Terrain System defines different terrains, which contains different sources and climates. Buff System contains four main buffs, such as basic buff, environment buff, terrain buff and equipment buff. Event Handling System gathers all the needed information from Biological Attribute System, Item System, Terrain System, and Buff System, and take effects in these systems. After finishing the event, it reports the result of the action and new observations to Action System, which is further transferred to the RL algorithm.

\begin{figure}[htbp]
	\centering
	\includegraphics[width=0.5\textwidth]{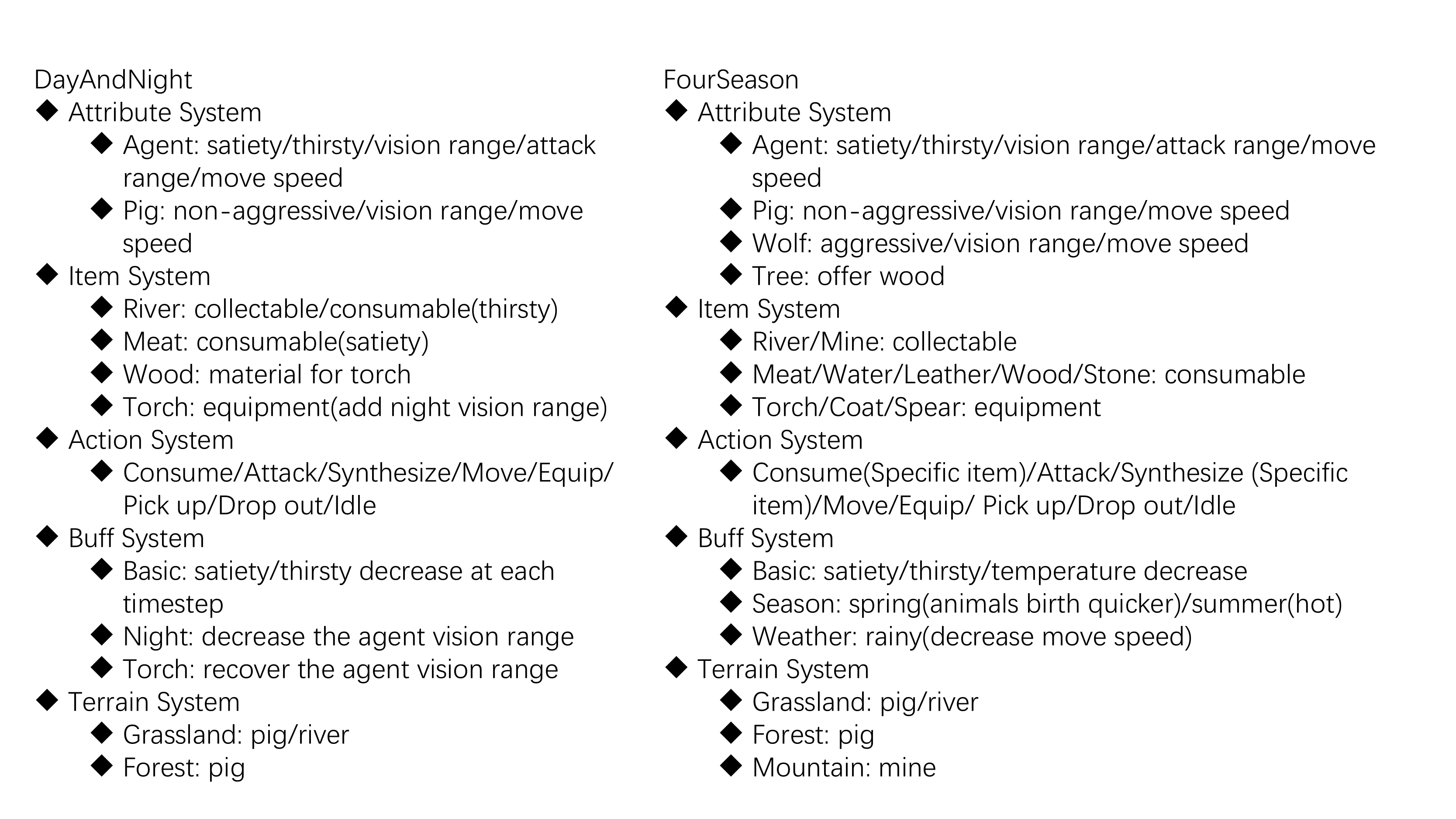}
	\caption{Two examples of ocnfigured environments: DayAndNight and FourSeason.}
	\label{fig:subsystem}
\end{figure}

In the “DayAndNight” world, we add four types of creatures as agent, pig, tree, and river, each of them having certain attributes in the Attribute System. Agent is the creature connected to the reinforcement learning algorithm, which has five attributes as shown in Figure~\ref{fig:subsystem}. Pig is a representative of the non-aggressive animals, which means a pig does not attack the agent and would run away if the agent moves close to it. Pig has two attributes as shown in Figure~\ref{fig:subsystem}. Pig can be attacked by the agent, and when it is killed, agent can get certain items. Tree and river are added so that the agent can collect them to get some other items. We use add four kinds of items as water, meat, wood, and torch in the Item System. When a river is collected by the agent, water will appear in the map; When a pig is killed by the agent, meat will appear in the map; When a tree is collected by the agent, wood will appear in the map. Water and meat can be consumed, and agent can gain thirsty and satiety from the consumption. Wood is the raw material of torch and can be used in the synthesis action. As described above, actions in this world can be chosen from consume, attack, synthesize, and move, etc., which makes up our Action System. The highlight of the world is Buff System, in which we get a buff (actually, it is a de-buff for the agent) brought by night, which reduces the vision range of the agent. Agent is expected to get a torch, which recovers the vision range with another buff, to overcome this obstacle. When creating such a world, Terrain System firstly decides the geography type of an area, and then certain types of creature or items are more likely to appear in this area.

The “FourSeason” world adds more complexity in the previous “DayAndNight” world. Unlike the pig, wolf is a representative of aggressive animals, which mean a wolf tries to attack the agent in its vision field. In order to deal with this dangerous animal, we allow the agent to equip itself with better equipment as coat (for defense) and spear (for attack). Also, season and weather are introduced to this world to make the world more challenging. Agent is expected to prepare for the upcoming season, which is harder to survive in, and deal with the changing weather. As is shown, new features can be easily added to an existing world with minor modification of the systems. New creatures can be added to the world by merely modifying Biological Attribute System following the template of pig, tree, or wolf, and the distribution of these creatures in the map should be configured in Terrain System. With these new creatures, users may need to gain new items with new buffs, and this can be realized by modifying Item System and Buff System. Moreover, users may also need new actions to deal with the new creatures and items, which can be easily realized by modifying Action System. Besides, all kinds of special effects relating to time or geography, like the examples of night or season, can be added to the world by modifying Buff System and Terrain System.

\subsection{Configuration Options}
\subsubsection{Multi-Dimension State}
\begin{figure}[htbp]
	\centering
	\includegraphics[width=0.5\textwidth]{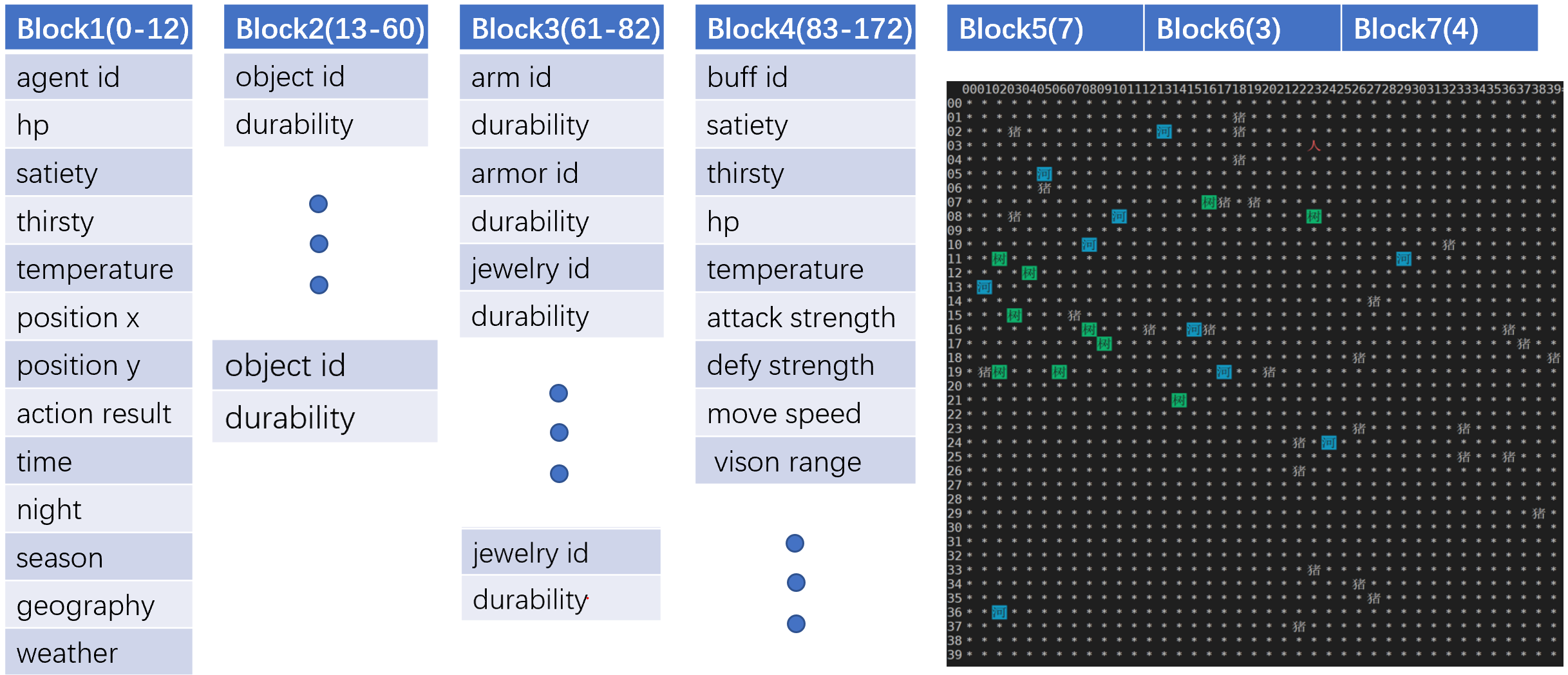}
	\caption{The structured input for the agent, and each element has different logic meanings.}
	\label{fig:jihuang_state}
\end{figure}

The agent state is a long vector, and each element of this vector has different meanings. The elements can be categorized into six blocks, as shown in Figure~\ref{fig:jihuang_state}. Block1 denotes the basic attribute of the agent, and the line of action result denotes whether the previous action is successful. Block2 denotes the objects in the backpack, and the backpack capacity is 24. Each object has two elements, the object id and its durability. Block3 denotes the agent equipment, and each equipment has an id and its durability. Block4 denotes the buff acted on the agent. In DayAndNight and FourSeason, there are nine different buffs in these environments. Block5-Block7 denote the objects in the agent vision. In Block5, the length of each object is 7. For example, if the object is an animal, each element of this 7-th vector means animal type, position x, position y, hp, and three reserved bits. In Block6, the length of each object is 3, and each element means type, position x and position y. Each object means one material in the map, and the material can be stacked. In Block7, the vector length is 4, and denotes the whether and geography of the position in the map.

\begin{figure}[htbp]
	\centering
	\includegraphics[width=0.5\textwidth]{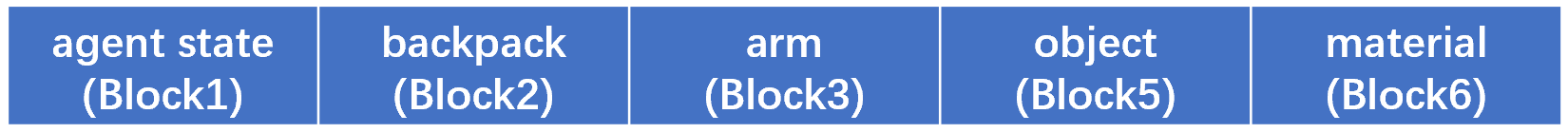}
	\caption{The logic meaning of each element of multi-dimension state.}
	\label{fig:multi_state}
\end{figure}

The dimension of the original state is too large, which is not conducive for the agent to extract key information. Therefore, different dimensions of the state are designed to help the agent accelerate the speed of information process. We delete some unrelated information that has little effects on agent survival to form the multi-dimension state, as shown in Figure~\ref{fig:multi_state}. The designed state contains five different kinds of information. Take the DayAndNight environment as an example, the agent state block denotes the agent status from Block1 in Figure~\ref{fig:jihuang_state}. The backpack block is reorganized according the number of items in the backpack, such as the first element means the meat number in the backpack, the second element means the water number and so on. As there are only one kinds of arm, Torch, in the DayAndNight environment, the arm block only means whether the torch is equipped by the agent. The object block and the material block denotes the objects and the materials that the agent can observe in the map. The multi-dimension state in this paper means how many objects that the agent can observe in the map. If the agent can only see the nearest objects and materials, the designed state is 78-dimensional vector. If the agent can see the 10 closest pigs, it is an 105-dimensional vector. Moreover, if the agent can see the 10 closest objects, the dimension of input vector is 195. If the original vector is kept, the vector dimension is 4203.

\begin{figure*}[htbp]
	\centering
	\includegraphics[width=16cm,height=9.0cm]{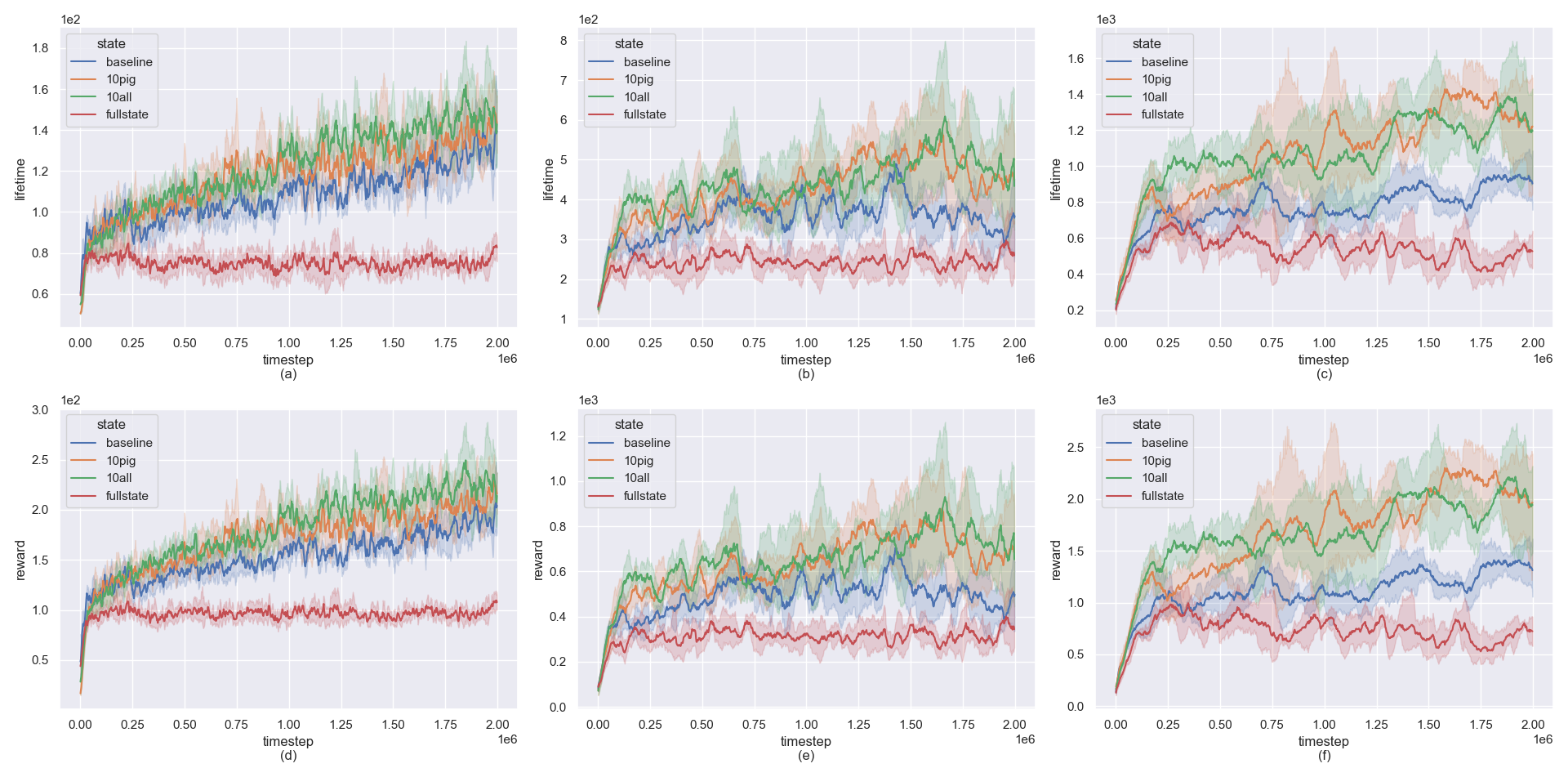}
	\caption{The experiments of multi-dimension sate. The upper part denotes the agent lifetime, and the lower part denotes the agent rewards, where the life-limit for (a) and (d) is 50, the life-limit of (b) and (e) is 100, and the life-limit of (c) and (f) is 150.}
	\label{fig:multi_state_exp}
\end{figure*}

Some experiments of multi-dimension state are shown in Figure~\ref{fig:multi_state_exp}. In all these experiments, agents are trained with PPO using dense reward function, which is further introduced later in the Reward section. In Figure~\ref{fig:multi_state_exp}(a)(d), the life-limit is set to 50, which means that the upper limit of satiety and thirsty volume is 50. The baseline agent achieves better performance at the beginning of training process, yet the curve is volatile, meaning that the training process is highly random. The reason may be that the second closest pig would become the closest one, and then the pig position will change dramatically. Due to this reason, episodes with extremely long or short lifetime may appear during the training process, leading to higher variance in lifetime. Meanwhile, the training difficulty of the baseline agent is lowest, then the baseline agent converge quickly. Moreover, the agent with 195-dimensional state vector(the 10all curve) achieves the best performance at the end of training process. In Figure~\ref{fig:multi_state_exp}(b)(d), the life-limit is set to 100. It can be found in the figure that baseline PPO agent can achieve better performance at the very beginning of training process, and the variance at this stage is relatively small. Meanwhile, the PPO agents with the 105-dimensional state vector(the 10pigs curve) and the 195-dimensional state vector(the 10all curve) achieve similar results with this configuration. The reason may be that the agent with lower life-limit should pay much more attention to killing pigs as this task is harder than collecting water. Therefore, other objects take little effects on the agent objective of living longer. In conclusion, the more information in the input state vector, the higher performance PPO agent can achieve. Besides, higher dimensional information makes it harder to train an agent.

\subsubsection{Multi-Dimension Action}
\begin{figure}[htbp]
	\centering
	\includegraphics[width=0.5\textwidth]{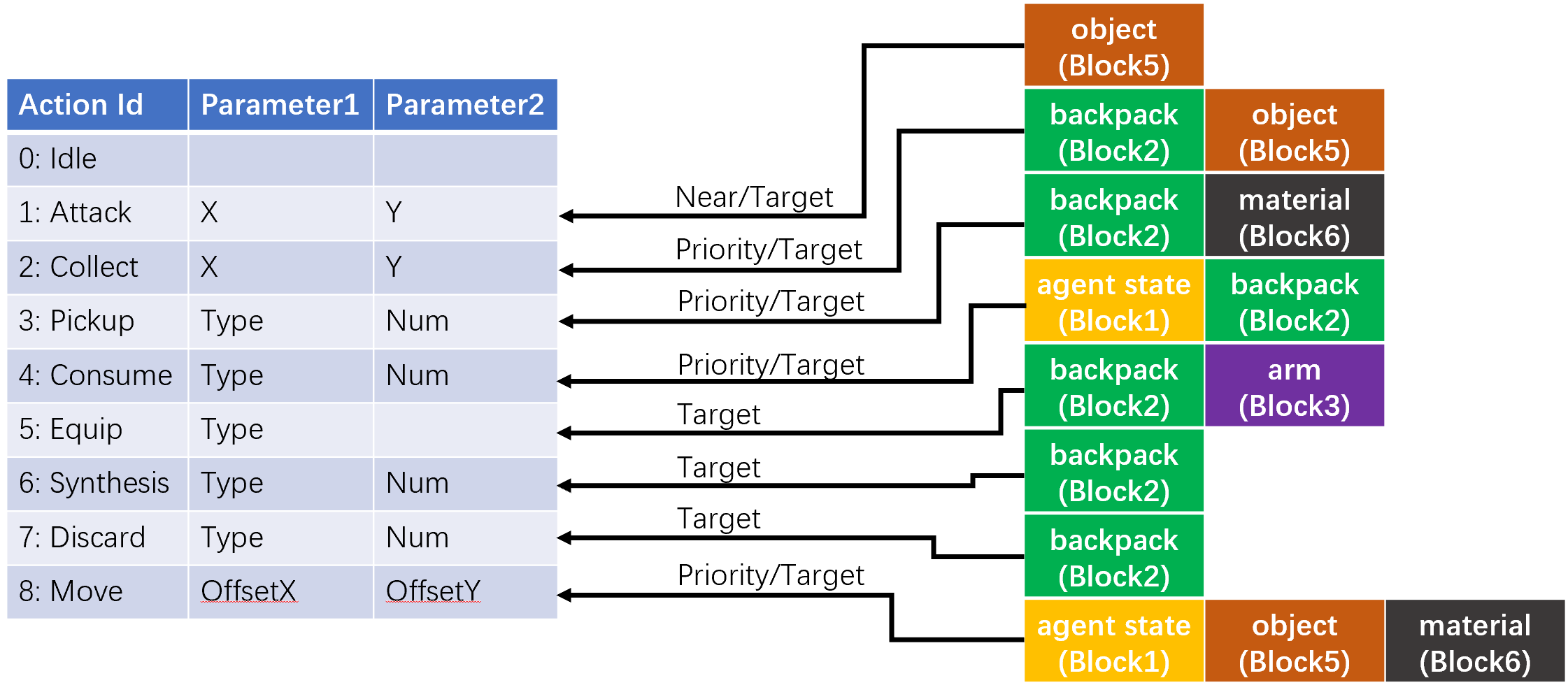}
	\caption{The action space of Eden. The left part denotes the original action space, and the right part denotes the simplified action space provided by the environment.}
	\label{fig:jihuang_action}
\end{figure}

The action space of the agent is shown in Figure~\ref{fig:jihuang_action}. Each action has three parameters, the action type id and the action parameters. If the action id is attack or collect, the action parameters mean the object position in the map. If the action id is pick, the first parameter means the picked materials, and the second parameter denotes the material count. If the action id is consume, synthesis or discard, the first parameter denotes the object type, and the second parameter means the object count. If the action id is equip or unequip, only the first parameter is valid and means the object type. When the equipment is the same as the first parameter and there is no equipment in the corresponding equipment position, the agent would equip the equipment. When the corresponding equipment position has one equipment, the agent would unequip the equipment. If the action id is move, the parameters mean the relative displacement increment $OffsetX$ and $OffsetY$. When the current agent position is $(x,y)$, the next agent position will be $(x+OffsetX,y+OffsetY)$.

\begin{figure*}[htbp]
	\centering
	\includegraphics[width=0.8\textwidth]{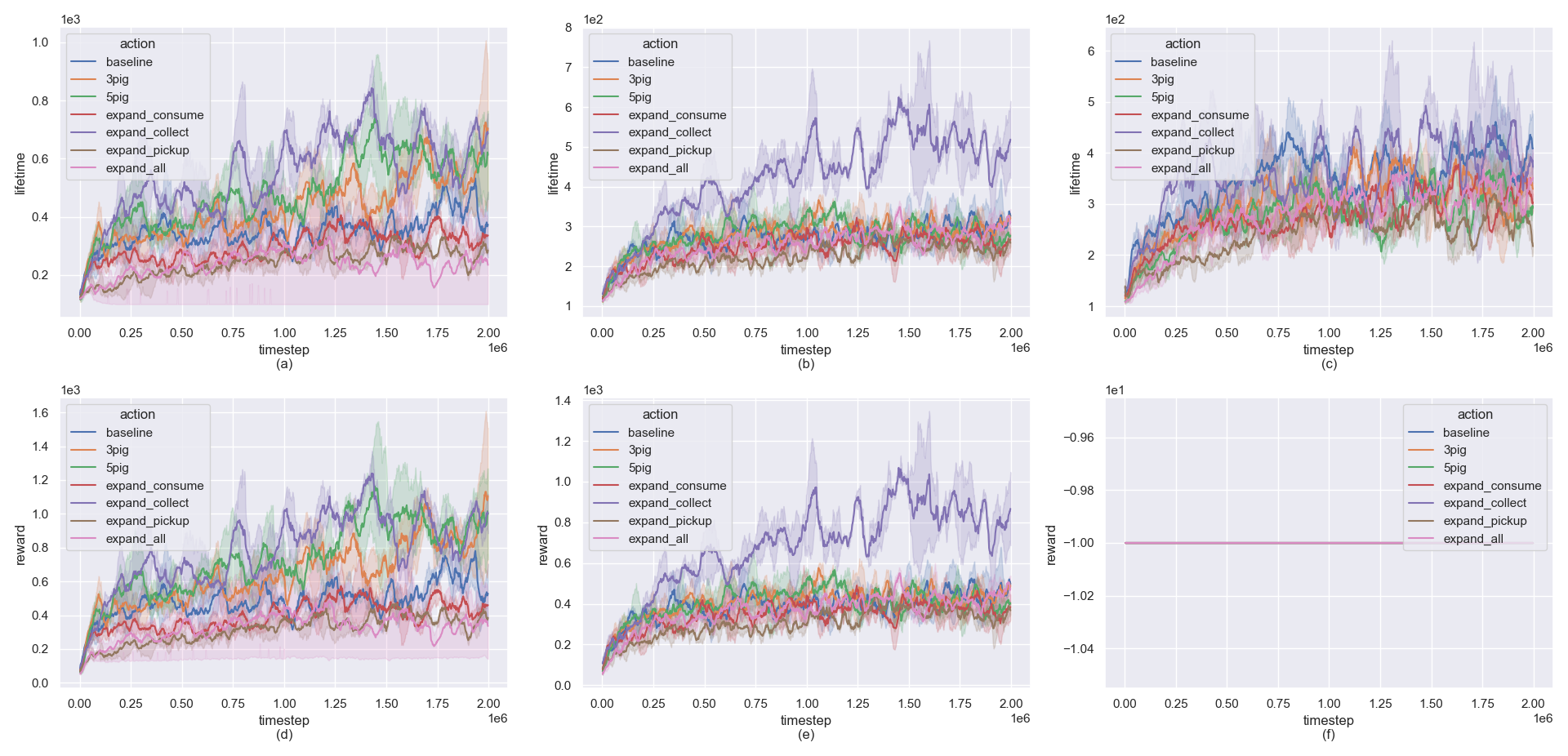}
	\caption{The experiments of multi-dimension action. The upper part denotes the agent lifetime, and the lower part denotes the agent total rewards. Experiments in (a)(d) are trained with PPO using dense reward, while (b)(d) using sparse, and (c)(f) using very sparse. }
	\label{fig:action_exp}
\end{figure*}

To simplify the training process, we make some simplifications of the original action space. For the baseline agent, the action space is simplified into 9 discrete actions. In summary, the agent would attack the nearest prey in the vision range for the attack action; the agent would choose to collect river or tree based on the amount of water and wood in the backpack for the collect action; the agent would choose to pick up meat, water or wood based on the amount of materials in the backpack for the pickup action; the agent would consume meat or water based on the agent satiety and thirsty value for the consume action; the first parameter is the type id of torch for the equip, synthesis and discard actions; the agent would move to pigs if there is any pig in the agent vision for the move action. As shown in Figure~\ref{fig:action_exp}, we also expand these 9 discrete actions to extend the performance limit of our simplified actions. The life-limit of agent is set to 300, and the agents are trained with PPO using dense((a)(d)), sparse((b)(e)), and very sparse((c)(f)) reward function respectively. In summary, the 3pig and the 5pig mean that the agent can decide to attack which pig in its vision range; the expand\_consume means that the agent can decide whether to consume meat or water; the expand\_pickup means that the agent could decide whether to pick up meat, water or wood; the expand\_all means to expand all the actions above.

With these experiments, we find some interesting phenomenon. First, agents tend to perform better with a few more choices, but when we present too many choices their performance becomes not satisfying. The reason for not performing well with large amount of choices mainly lie on 
the exponential increase of action combinations when each element in the combination has too many choices. However, with a few more choices added, agents are able to specify the targets of their action and thus find some useful combinations, instead of being restricted by the auto target choice of tasks. Second, expand\_collect experiments outperform the others with sparse reward function, while with the dense or very sparse ones these experiments do not have significant differences. Reason for this promotion lie on the sparse reward design, which returns rewards each time the agent accomplish a consuming task. The expand of collect actually offers a choice for the agent to fully neglect the task of getting a torch, which originally shares the collect action with the task of getting water. Along with the sparse rewards, this makes it easier for the agent to accomplish the tasks of consuming water or meat, thus leading to a local optimum, which outperforms the others remarkably. Third, all the experiments achieve a reasonable performance with dense reward function, except for the expand\_all and the expand\_pickup experiments. The reason, again, may lie on our handcrafted dense reward design, which may mislead the algorithm into a local optimum worse than common results. From all these, we can see it that the action space has a complex influence on agents' performance, and this influence is closely related to reward design as well.

\subsubsection{Reward Function}
\begin{figure*}[htbp]
	\centering
	\includegraphics[width=0.8\textwidth]{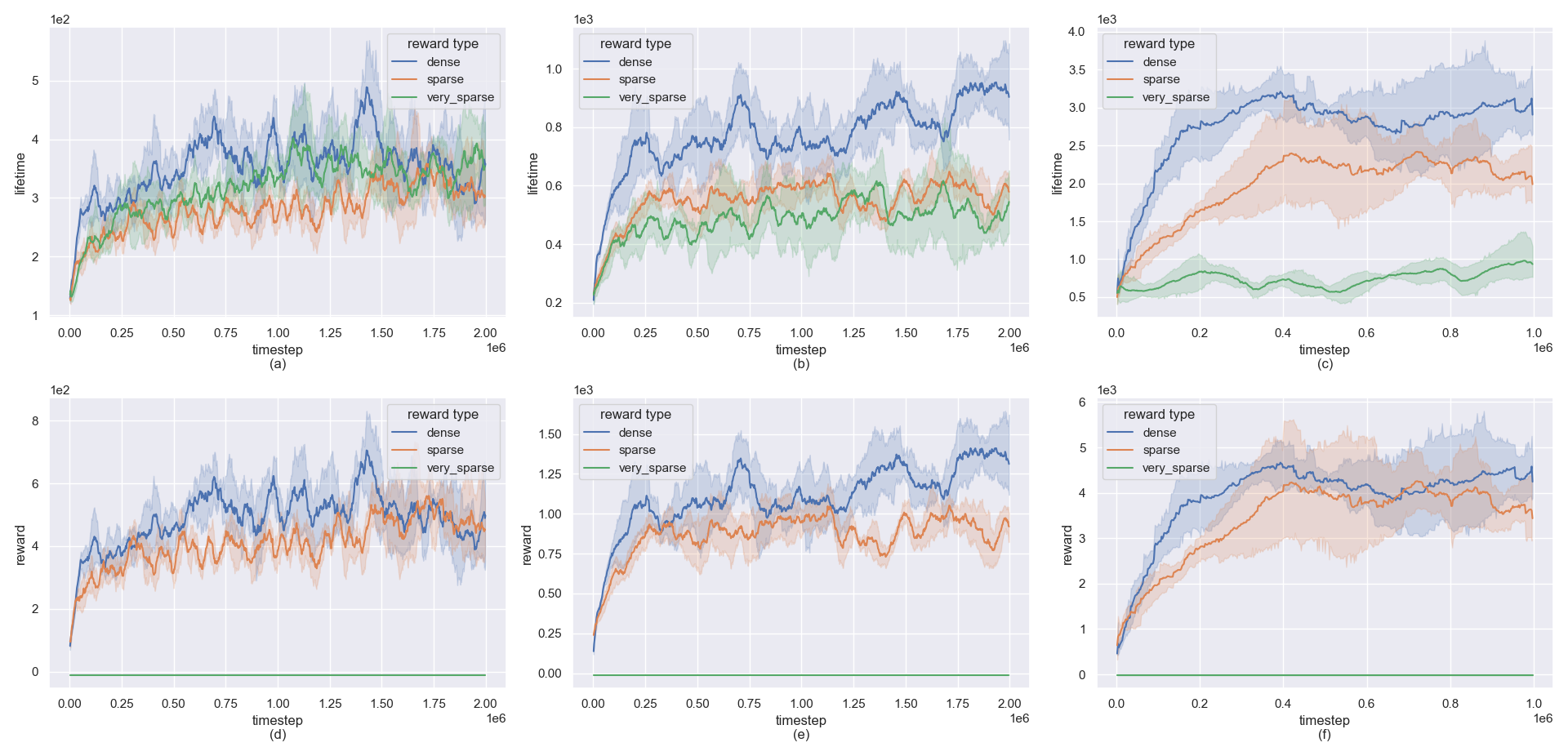}
	\caption{The experiments of three reward functions used for training. All the agents are trained with PPO algorithm. (a)(d) sets life-limit to 100, while (b)(e) set to 150, (c)(f) set to 300.}
	\label{fig:reward_exp}
\end{figure*}

\begin{figure}[htbp]
	\centering
	\includegraphics[width=0.5\textwidth]{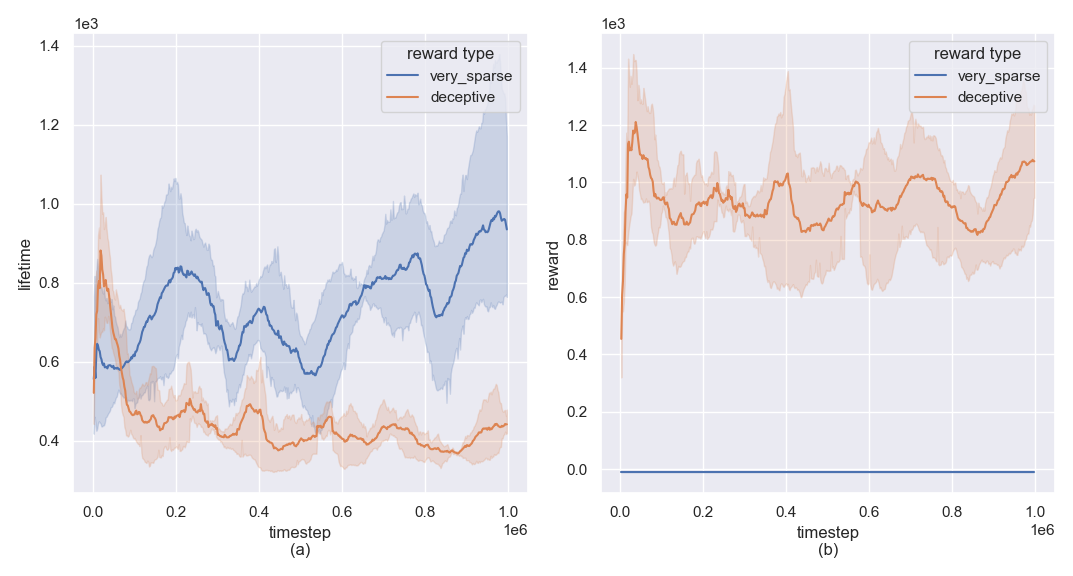}
	\caption{The experiments of deceptive reward function. The agents with life-limit set to 300 are trained with PPO algorithm, using very sparse and deceptive reward function respectively.}
	\label{fig:deceptive_reward}
\end{figure}

Considering the compatibility between our environment and main RL algorithms, we propose three different reward functions in our environment, namely dense, sparse and very sparse. Dense reward function rewards every step of the agent, meaning that we set a reward value for each action according to current state. Sparse reward function returns rewards only when certain tasks related directly to its survival are accomplished, and in our settings these tasks are consuming meat or water namely. Very sparse reward function only gives a fixed negative reward at the end of an episode.
Taking PPO as a representative of model-free algorithms, we present several experiments with different settings. The results can be viewed in Figure~\ref{fig:reward_exp}. In Figure~\ref{fig:reward_exp}(a)(d), life-limit is set to 100. In this setting, we can see a little difference among the results of three reward functions. In Figure~\ref{fig:reward_exp}(b)(e), life-limit is set to 150. We can see that dense reward function outperforms the other two. In Figure~\ref{fig:reward_exp}(c)(f), life-limit is set to 300, and we can see a remarkable difference among the results of three reward functions. From the experiments, we can draw the conclusion that dense reward function is usually a good choice for model-free algorithms, especially for high life-limits.
Moreover, we modify some of the return values in dense reward function to form a deceptive reward function. With this change, we present a few experiments in Figure~\ref{fig:deceptive_reward}, and we compare them to those trained using very sparse reward function. As discussed before, usually we expect a better performance when we use the dense reward function, yet we can see that agents fail to live long when trained with this deceptive reward function. The lifetime of agents become to decrease quickly after a few episodes. However, we can only see a small decrease in total rewards, meaning that the algorithm are taking effect, but the performance can not be improved by following the instruction of reward function. This experiment shows that we must be careful when designing a new reward function, otherwise it may not help with the training process.

\subsubsection{Survival Stress}

\begin{figure*}[htbp]
	\centering
	\includegraphics[width=0.9\textwidth]{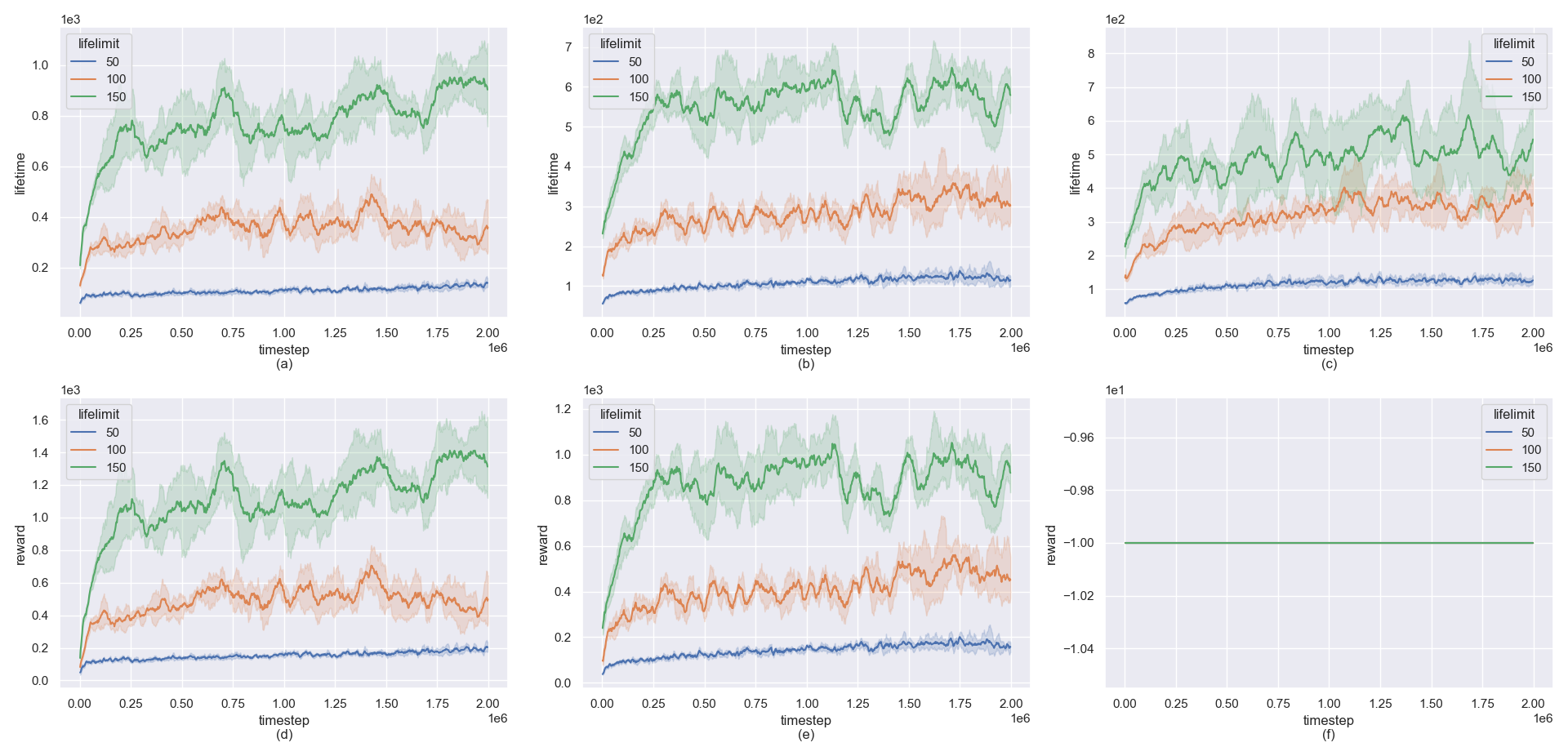}
	\caption{The experiments of survival stress. The agents are trained with PPO with life-limit set to 50, 100, 150 respectively. Experiments in (a)(d) are trained with dense reward function, (b)(e) with sparse, and (c)(f) with very sparse reward.}
	\label{fig:survival_stress}
\end{figure*}

In this paper, survival stress of an agent is determined by its upper limit of satiety and thirsty (usually, these two are set to the same value), which is mentioned as life-limit. Generally, higher value of life-limit leads to smaller survival stress. As discussed in Task Difficulty section, we know it that higher life-limit makes it easier for the agent to accomplish the goal set for survival, and therefore easier to survive. While the average performance can be improved, higher life-limit further makes it more possible for agents to suffer from deceptive trajectories, which have long lifetime, yet the decision during this trajectory may not be reasonable all the time. As a result, agents with higher life-limit sometimes perform unstably. \\
We present a series of experiments under the three previously discussed reward designs and different life-limit settings. In Figure~\ref{fig:survival_stress}, we present experiments under three different reward designs and three life-limit settings, all using PPO algorithm. Figure(a)(d) shows the mean length of episodes (i.e. lifetime) and mean rewards per episode under the dense reward function. As we can see, agents with higher life-limit have remarkably longer lifetime in this settings, while its performance may be more volatile. Figure(b)(e) are performed under the sparse reward function. In general, the trend of there curves are very much like the one under a dense reward, but the values of life-limit and rewards are a little smaller.  Figure(c)(f) are performed under the very sparse reward function. In this setting, we can see that the curve of life-limit 150 is highly volatile, while the general numerical relationship between different life-limit setting remain similar to the previous settings.\\
In summary, when we takes a comprehensive view of the three experiments, we can see the following features: 1. Higher life-limits lead to higher lifetime. 2. Higher life-limits may lead to more volatile performance.

\subsection{Evaluation}
\subsubsection{Task Time Evaluation}
We firstly introduce a metric calling Task Time Maximum(TTMX), which is unrelated to algorithm or reward design, for calculating maximum time required for some task in our environment, so is Task Time Minimum(TTMN) for the minimum time. Intuitively, a task is more difficult when more types of action are required for accomplishing it, and especially difficult when the action are required to be repeated and arranged in a certain order. We formally introduce TTMX as a metric to evaluate such difficulty in the arrangement of actions.
For an episode lasting for $N$ steps, we have an array of action and state. We set $\tau^N=((a_1, s_2), (a_2, s_2)...)$ as the trajectory of this episode, action and step at time $i$ is mentioned as $\tau_i^N$. 
The TTMX of a certain task is defined as the minimum number of random actions that makes the task complete: 
\begin{align}
TTMX(\mathcal{T})\ =\  \min_{t}{\left\{t\in\mathbb{N}\middle|\sum_{N=1}^{t}\frac{1}{\#\{\tau^N\}}\sum_{\tau^N}\prod_{i=0}^{N}p(a_i) \ge th\right\}}, \label{TTMX_definition} \\
\end{align}
$\forall \tau^N$ which, $\tau_N^N$'s state satisfies the task, and $\tau_j^N$'s state, $\forall j\in [0, N-1]$ do not satisfy the task, i.e. firstly satisfies the task at time step $N$. $th$ is a threshold, higher $th$ leads to higher value of a metric. In practice, we recommend a value satisfied $0.9\le th < 1$ . $\mathcal{T}$ is a task in our environment. meanwhile, the TTMN is defined as the minimum number of actions that makes the task complete, i.e.  $TTMN(\mathcal{T})\ =\ \min_{t}{\left\{t\in\mathbb{N}\middle|\tau^N \text{can complete the task}\right\}}$. The interval $(TTMN(\mathcal{T}),\ TTMX(\mathcal{T}))$ represents the area of time required to complete the task. \\
To estimate the TTMX without traversing the whole space of trajectory, we propose to consider the average least steps to accomplish the task $\mathbb{E}(t_{\mathcal{T}})$, and use deterministic policy to approximate the probability of accomplishing it at any step. Meanwhile, $\mathbb{E}(t_{\mathcal{T}})$ is also an estimate of $TTMN(\mathcal{T})$. The average least steps are estimated using two factors:
\begin{align}
\widehat{TTMN}(\mathcal{T})\ =\ \mathbb{E}(t_{\mathcal{T}}) = f(\mathbf{g}(\mathcal{T}), \mathbf{e}(\mathcal{T})), 
\end{align}
$\mathbf{g}(\mathcal{T})$ is the cardinal number of the action set except "Move" required to accomplish the task. $\mathbf{e}(\mathcal{T})$ is an environment-determined parameter, which measures how many times should a certain type of action be repeated on average, i.e. the timesteps spent on moving. A common function $f$ just adds $\mathbf{g}(\mathcal{T})$ and $ \mathbf{e}(\mathcal{T})$. For simplicity, we denote $\mathbb{E}$ as the smallest integer larger than $\mathbb{E}(t_{\mathcal{T}})$. Then, an estimation for the left part of inequality Eq.\eqref{TTMX_definition} is:
\begin{align}
p_T(t)= \left\{\begin{matrix}
\sum\limits_{\Sigma n_i = t-\mathbb{E}, \atop n_i \in \mathbb{N}, i=1,...,\mathbb{E}}\prod\limits_{j=1}^{\mathbb{E}}p_{j}^{n_{j}}p_{j}^*,\text{if }t \geq \mathbb{E}
\\
\prod\limits_{j=1}^{t}p_{j}^*, \text{if }t < \mathbb{E}
\end{matrix}\right. ,
\label{PTT_definition}
\end{align}
where $p_j$ is the probability of a valid but useless action being selected during time step $n_{j-1}$ to $n_j$ and $p_j^*$ is the probability of a certain type action, which is exactly the one contributing to accomplishment of the task, to be selected.
\\
With this assumption, we can estimate the distribution of time step at which agent accomplishes the task for the first time:
Combined with Eq.\eqref{PTT_definition}, we get an estimate of the minimum integer solution satisfying inequality Eq.\eqref{TTMX_definition}:
\begin{equation}
\widehat{TTMX}(\mathcal{T})\ =\ \min_{x}{\left\{x\in\mathbb{N}\middle|p_T(t)\ge th\right\}}
\end{equation}
Thus, an estimate of the time interval $(TTMN(\mathcal{T}),\ TTMX(\mathcal{T}))$ is  $[\mathbb{E}(t_{\mathcal{T}_{goal}}),\ TID(\mathcal{T})]$. \\
\\
\subsubsection{Goal Difficulty Level}
For completing one task $\mathcal{T}$, we can get an estimate of time interval, $[\mathbb{E}(t_{\mathcal{T}_{goal}}),\ TID(\mathcal{T})]$, through the above calculation. $\forall n\in \{2,\ 3,\ ...,\ TID(\mathcal{T})-\mathbb{E}(t_{\mathcal{T}_{goal}})\}$, $\forall {\{a_m\}}_{m=0}^n$ satisfying $\mathbb{E}(t_{\mathcal{T}_{goal}})=a_0<\ a_1<\ ...<a_m=TID(\mathcal{T}),\ a_j \in\mathbb{N}\ for\ m=0,1,...,n$. We divide the interval $[\mathbb{E}(t_{\mathcal{T}_{goal}}),\ TID(\mathcal{T})]$ into n parts, $[a_i,\ a_{i+1}]\ for\ i=0,1,...,n-1$. Thus we can define goal from a task:` Complete the task before $a_i$-th timestep', concurrently goal difficulty level is defined as the $'i+1'$, so there $n$ levels for a task.
\\
To quantitatively evaluate the difficulty of a environment-agnostic and algorithm-agnostic task, \cite{furuta2021policy} recently proposed \emph{policy information capacity (PIC)}. They defined PIC as a mutual information:
\begin{equation}
\mathcal{I}(R;\Theta)=\mathcal{H}(R)-\mathbb{E}_{p(\theta)}[\mathcal{H}(R|\Theta=\theta)],
\label{PIC_definition}
\end{equation}
where $\mathcal{H}(\cdot)$ is  Shannon entropy, $R$ is the cumulative reward, $\Theta$ is the prior policy parameter. We can estimate Eq.\eqref{PIC_definition} via discretization of the empirical cumulative reward distribution for $p(r)$ and each $p(r|{\theta}_{i})$ using the same $B$ bins. They set min and max values observed in sampling as the limit, and divide it into $B(>M)$ equal parts( not similar to the task time divide):
\begin{align}
\hat{\mathcal{I}}(R;\Theta)=-\sum_{b=1}^{B}\hat{p}(r_b)log\hat{p}(r_b)\notag \\
+\frac{1}{N}\sum_{i=1}^{N}\sum_{b=1}^{B}\hat{p}(r_b|{\theta}_i)log\hat{p}(r_b|{\theta}_i)
\end{align}
where $p(r)=\mathbb{E}_{p(\tau|\theta)p(\theta)}[p(r|\tau)],p(r|\theta)=\mathbb{E}_{p(\tau|\theta)}[p(r|\tau)]$, $p(\tau|\theta)=p(s_1)\prod\limits_{t=1}^{T}p({s}_{t+1}|s_t,a_t)\pi(a_t|s_t,\theta)$, $p(r|\tau)$ is the reward distribution over trajectory.
So when we set different upper time limits, reward distributions will adjust accordingly, and then we can calculation the PIC to evaluate the task difficulty to validate our evaluation method 'goal level'.

\section{Examples of RL Application}
In this section, we have obtained experimental results in the field of reinforcement learning such as Model-Free RL, Exploration Strategy, and Meta-RL by configuring the environment framework. These experimental results are similar to those described in the original papers, which demonstrates that this environmental framework can be applied to a variety of different reinforcement learning algorithms.

\subsection{Model-Free RL}
\begin{figure}[htbp]
	\centering
	\includegraphics[width=0.8\textwidth]{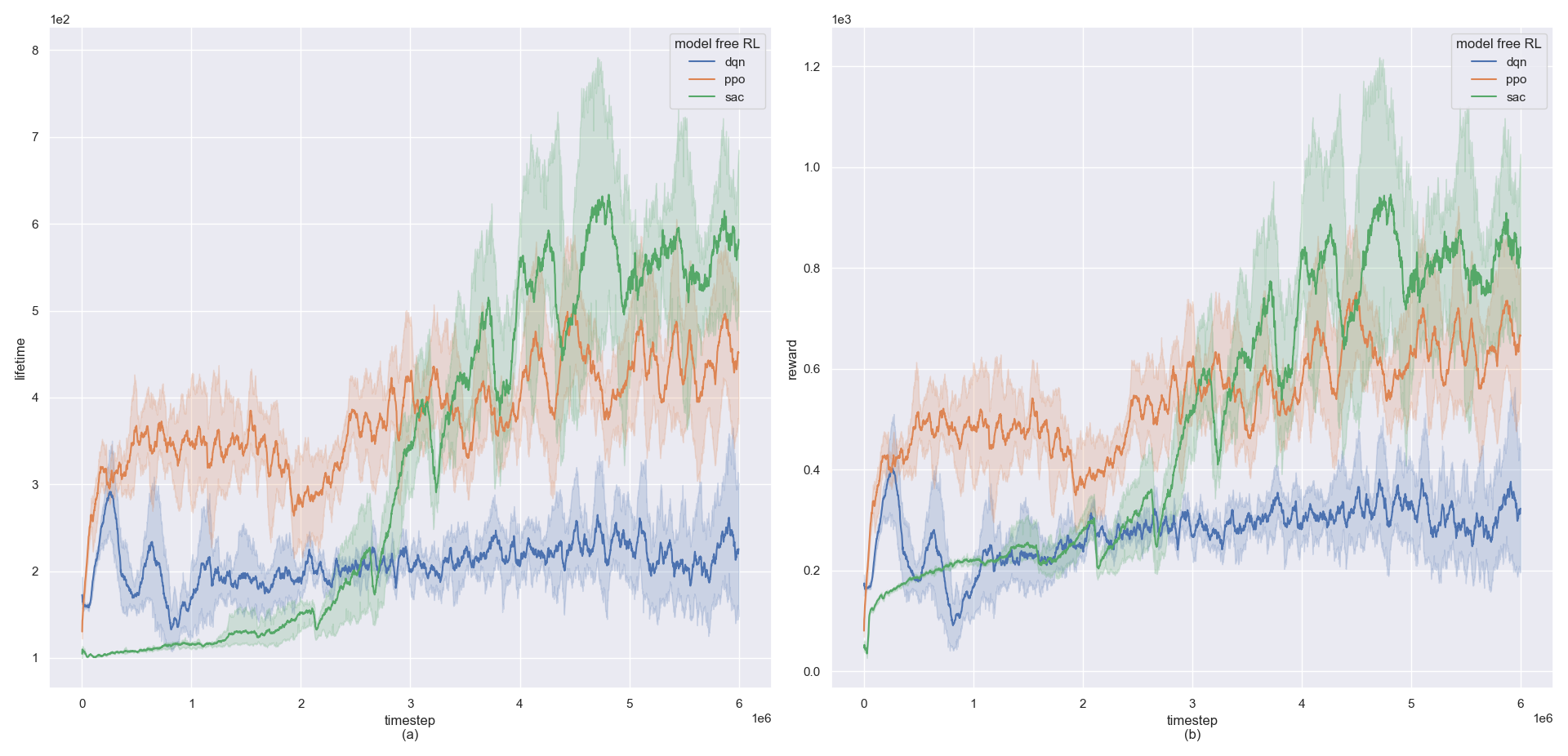}
	\caption{Model Free Experiments}
	\label{fig:model_free_exp}
\end{figure}

All the experiments in Figure ~\ref{fig:model_free_exp} are performed using the configuration that upper limits of satiety and thirsty are set to 100. This configuration gives medium survival stress.a

Deep-Q-Learning (DQN) is a value-based model-free algorithm. In our training progress, the parameter $\epsilon$ in $\epsilon$-greedy update method is set to 0.9 at the very beginning, so that the agent can fully explore the environment at this stage. During this stage, we can see a little rise and fall in both lifetime and rewards as shown in Figure ~\ref{fig:model_free_exp}, since the algorithm are trying different actions, which leads to random behaviors. After full exploration, the parameter $\epsilon$ gradually decreases, leading to a stable stage since the policy has less randomness, following which we can see an increasing stage, meaning that this algorithm works correctly in our environment.

Soft-Actor-Critic (SAC) is an off-policy actor-critic algorithm, which takes entropy as an optimization objective. In our training progress, we manually adjust the parameter $\alpha$ in the optimization objective. We initialize $\alpha$ with 0.2, and doubles every 400,000 steps until it reaches 25.6, and then keep this value in the rest of training procedure. As shown in Figure ~\ref{fig:model_free_exp}, life-steps and reward are rather small in the beginning, because the entropy at this stage is high and the algorithm pays little attention to it, meaning that the policy has high randomness. Similar to DQN, this stage helps with afterwards training procedure, and we can see that with the increase of $\alpha$, the algorithm gains high performance breakthrough in both lifetime and rewards.

\subsection{Exploration Strategy}

\begin{figure}[htbp]
	\centering
	\includegraphics[width=0.8\textwidth]{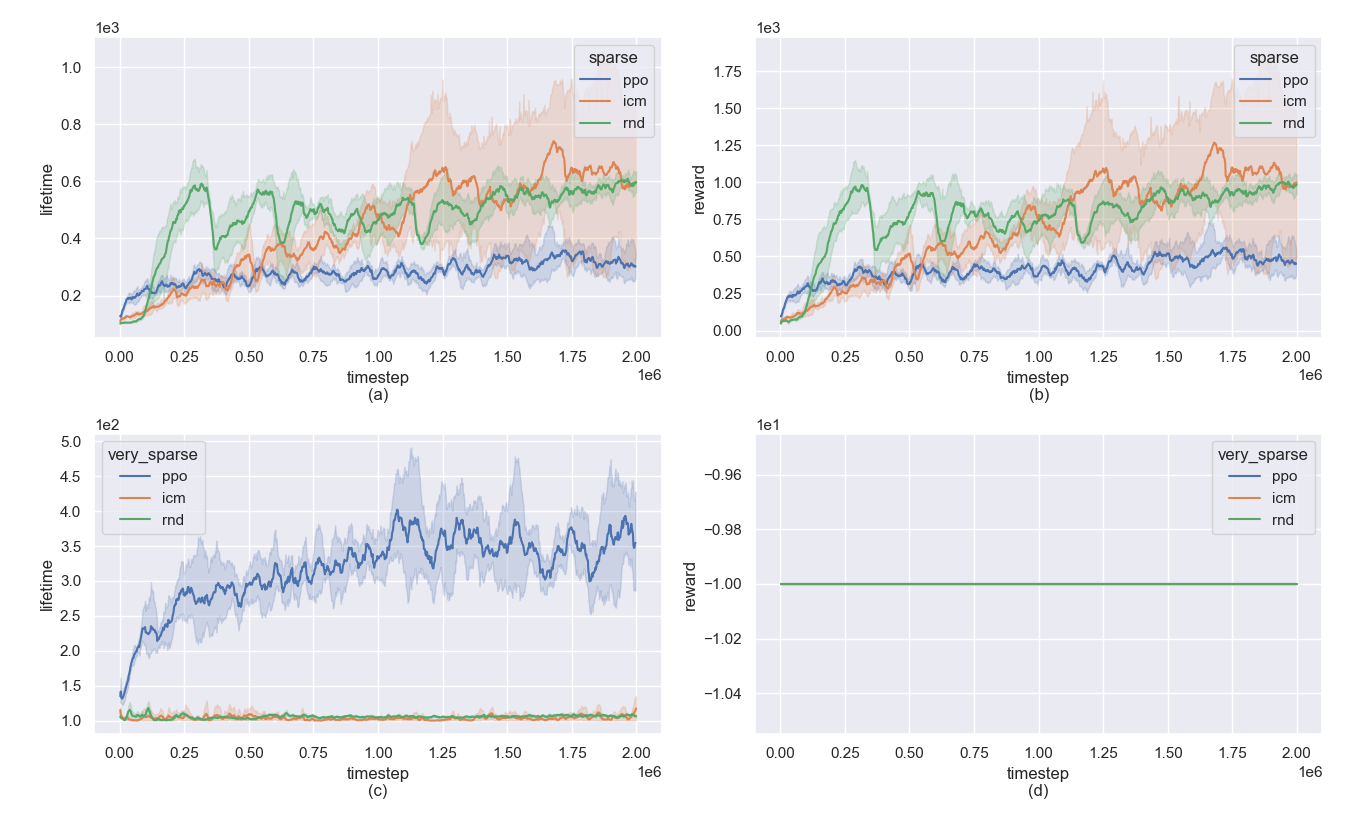}
	\caption{Exploration Strategy Experiments}
	\label{fig:explore_exp}
\end{figure}

Random-Network-Distillation (RND)  take use of exploration bonus which is able to combine with any policy optimization algorithm. While Intrinsic-Curiosity-Module (ICM) generates intrinsic inspiration via the explored states. In our experiments, we respectively combine them with PPO as the RL agent. The experiment results are shown in Figure~\ref{fig:explore_exp}.

When the reward is sparse, RND agent and ICM agent survive a longer lifetime than PPO agent, with the reward higher in the meaning time. However, RND curve is more volatile, while the ICM curve has a larger variance.  This might stems from the randomness of the exploration algorithms, which leads to an unavoidable fact that the exploration algorithm requires diversified information and valuable states while modeling the environment and generating efficient intrinsic motivation. In very sparse reward setting, RND and ICM are poles apart with PPO when both Meat and Water reward are absent. We provide a possible explanation to this phenomenon. 

Consider an initial state s. The agent can reach an arbitrary state t through several state transitions starting from s. All possible t constitute a state transition graph $G_{env}(s)$, and the radius of the graph starting from s is $r_{env}(s)$. Similarly, we can obtain a state transition graph $G_{explore}(s)$ with radius $r_{explore}(s)$ via an exploration algorithm starting from s. Assuming that an RL agent is supposed to traverse a state transition graph $G_{agent}(s)$ whose radius is $r_{agent}(s)$, to finish a certain environmental task well. When $r_{agent}(s)$> $r_{explore}(s)$, the exploration algorithm cannot cover the necessary states, resulting in agent's failure to learn a strategy to finish the task well. 

In very sparse reward setting, $r_{agent}(s)$ is greater than $r_{explore}(s)$ of RND and ICM, therefore both of them are defeated by PPO, while PPO is able to imperfectly finish the task in this setting. In sparse reward setting, $r_{agent}(s)$ is smaller than that in very sparse setting, even smaller than $r_{explore}(s)$ of RND and PPO, thus making both of them outperform PPO.

\subsection{Meta-RL}

In Eden-navigation-40x40, the agent must move to target river spots in a 2D grid environment. The size of the environment is 40x40, it has nothing but random generalized rivers. The locations of all water sources are randomly generated according to the seed given to the environment. The observation is the current location of the agent, and the action is the offset in both X and Y directions. We stipulate that to the right and the up is the positive direction. The offset is auto clipped into integers within range [-2, 2] in the back-end system of the environment. The reward is the  difference between last and current squared distance to the target river spot, episode terminates when the agent is within 0.1 of the goal or at the horizon of H=20, we set the ending horizon to 20 because the agent initializes in the center of the map.

\begin{figure}[htbp]
	\centering
	\includegraphics[width=0.8\textwidth]{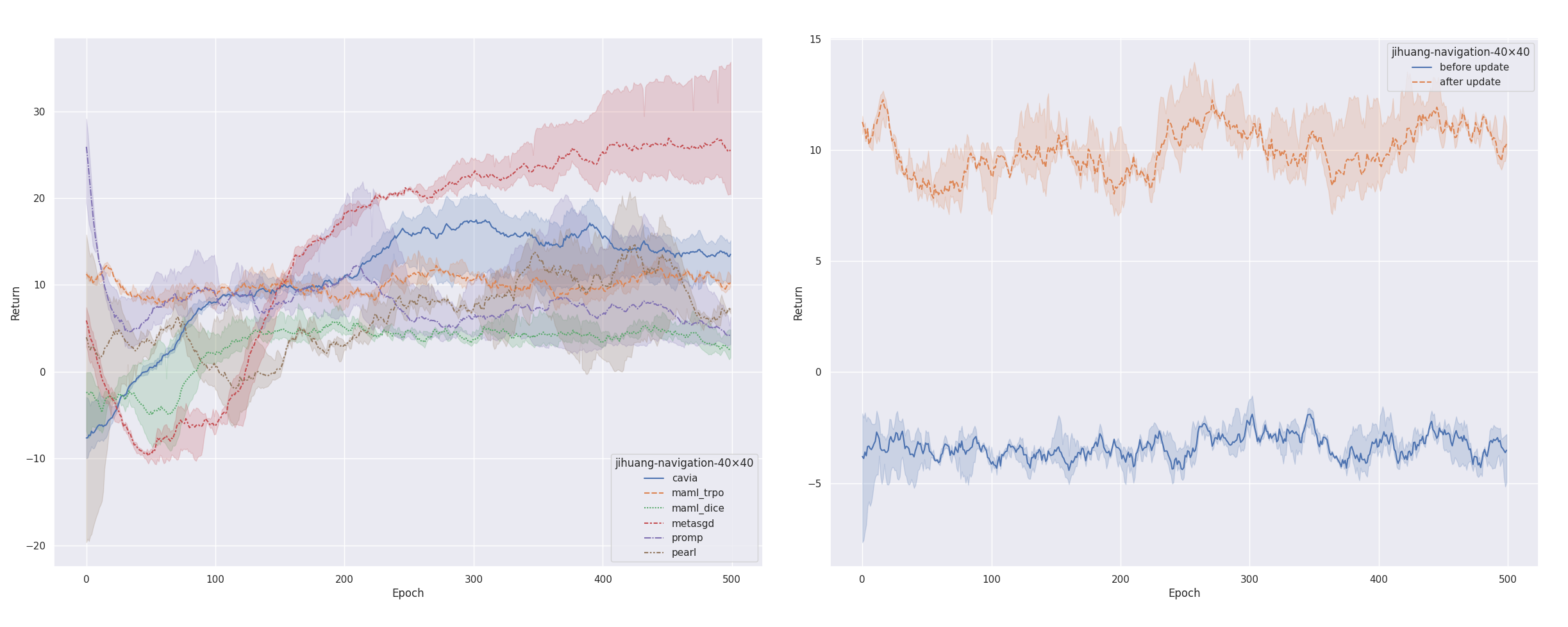}
	\caption{Meta-RL Experiments}
	\label{fig:meta-exp1}
\end{figure}

We test six optimization-based meta-rl algorithms in Eden-navigation-40x40: cavia, maml-trpo, maml-dice, metasgd, promp and pearl. The policy of each algorithm was trained with 1 policy gradient update using 20 episodes. Figure~\ref{fig:meta-exp1}(a) shows the update return of the algorithms over 500 epochs. The average update return of the algorithms is within range 0 to 20. Among them, maml dice performs worst and metasgd reaches highest return. Figure~\ref{fig:meta-exp1}(b) shows the return before update and after update of maml in our setting.

\section{Conclusion and Future Works}
In this paper, we propose an unified reinforcement learning environment framework Eden. This environment framework can be configured to obtain a variety of different instantiation environments, which can be conveniently used for testing various types of reinforcement learning algorithms. Testing different types of reinforcement learning algorithms under different configurations of the same environmental framework can more easily define the performance evaluation of different algorithms in different dimensions, thereby guiding the subsequent development of the algorithm field.

Up to now, we have completed the code of the environment framework, and tested the effectiveness of different instantiation environments using the PPO algorithm in various configurations. Meanwhile, we tested the performance of several reinforcement learning algorithms in the fields of Model-Free RL, Exploration Strategy and Meta-RL in the configured instantiation environment. In all those experiments, the algorithms in our instantiation environments achieve similar conclusions as the original papers.

In the future, we will continue to maintain and update this environment framework to meet more user requirements. We will further optimize the ease of use of the environment framework configuration and open source the code. At the same time, we will instantiate more environments to test more types of DRL algorithms. Finally, based on the above experimental results, we will design and implement an effective algorithm performance scoring system.

\bibliographystyle{unsrtnat}
\bibliography{references}  

\begin{thebibliography}{62}
\providecommand{\natexlab}[1]{#1}
\providecommand{\url}[1]{\texttt{#1}}
\expandafter\ifx\csname urlstyle\endcsname\relax
  \providecommand{\doi}[1]{doi: #1}\else
  \providecommand{\doi}{doi: \begingroup \urlstyle{rm}\Url}\fi

\bibitem[Silver et~al.(2017)Silver, Schrittwieser, Simonyan, Antonoglou, Huang,
  Guez, Hubert, Baker, Lai, and Bolton]{SilverSchrittwieser-67}
David Silver, Julian Schrittwieser, Karen Simonyan, Ioannis Antonoglou, Aja
  Huang, Arthur Guez, Thomas Hubert, Lucas Baker, Matthew Lai, and Adrian
  Bolton.
\newblock Mastering the game of go without human knowledge.
\newblock \emph{nature}, 550\penalty0 (7676):\penalty0 354--359, 2017.

\bibitem[Berner et~al.(2019)Berner, Brockman, Chan, Cheung, D{\k{e}}biak,
  Dennison, Farhi, Fischer, Hashme, Hesse, et~al.]{berner2019dota}
Christopher Berner, Greg Brockman, Brooke Chan, Vicki Cheung, Przemys{\l}aw
  D{\k{e}}biak, Christy Dennison, David Farhi, Quirin Fischer, Shariq Hashme,
  Chris Hesse, et~al.
\newblock Dota 2 with large scale deep reinforcement learning.
\newblock \emph{arXiv preprint arXiv:1912.06680}, 2019.

\bibitem[Yu et~al.(2020)Yu, Quillen, He, Julian, Hausman, Finn, and
  Levine]{YuQuillen-8}
Tianhe Yu, Deirdre Quillen, Zhanpeng He, Ryan Julian, Karol Hausman, Chelsea
  Finn, and Sergey Levine.
\newblock Meta-world: A benchmark and evaluation for multi-task and meta
  reinforcement learning.
\newblock In \emph{Conference on Robot Learning}, Conference on Robot Learning,
  pages 1094--1100. PMLR, 2020.

\bibitem[Zhang et~al.(2019)Zhang, Feng, Liu, Ding, Zhu, Zhou, Zhang, Yu, Jin,
  and Li]{ZhangFeng-86}
Huichu Zhang, Siyuan Feng, Chang Liu, Yaoyao Ding, Yichen Zhu, Zihan Zhou,
  Weinan Zhang, Yong Yu, Haiming Jin, and Zhenhui Li.
\newblock Cityflow: A multi-agent reinforcement learning environment for large
  scale city traffic scenario.
\newblock In \emph{The World Wide Web Conference}, The World Wide Web
  Conference, pages 3620--3624, 2019.

\bibitem[Team et~al.(2021)Team, Stooke, Mahajan, Barros, Deck, Bauer,
  Sygnowski, Trebacz, Jaderberg, Mathieu, et~al.]{team2021open}
Ended~Learning Team, Adam Stooke, Anuj Mahajan, Catarina Barros, Charlie Deck,
  Jakob Bauer, Jakub Sygnowski, Maja Trebacz, Max Jaderberg, Michael Mathieu,
  et~al.
\newblock Open-ended learning leads to generally capable agents.
\newblock \emph{arXiv preprint arXiv:2107.12808}, 2021.

\bibitem[Song et~al.(2020)Song, Wojcicki, Lukasiewicz, Wang, Aryan, Xu, Xu,
  Ding, and Wu]{SongWojcicki-85}
Yuhang Song, Andrzej Wojcicki, Thomas Lukasiewicz, Jianyi Wang, Abi Aryan,
  Zhenghua Xu, Mai Xu, Zihan Ding, and Lianlong Wu.
\newblock Arena: A general evaluation platform and building toolkit for
  multi-agent intelligence.
\newblock In \emph{Proceedings of the AAAI Conference on Artificial
  Intelligence}, volume~34 of \emph{Proceedings of the AAAI Conference on
  Artificial Intelligence}, pages 7253--7260, 2020.
\newblock 05.

\bibitem[Fujimoto et~al.(2019)Fujimoto, Conti, Ghavamzadeh, and
  Pineau]{fujimoto2019benchmarking}
Scott Fujimoto, Edoardo Conti, Mohammad Ghavamzadeh, and Joelle Pineau.
\newblock Benchmarking batch deep reinforcement learning algorithms.
\newblock \emph{arXiv preprint arXiv:1910.01708}, 2019.

\bibitem[McFarlane(2018)]{mcfarlane2018survey}
Roger McFarlane.
\newblock A survey of exploration strategies in reinforcement learning.
\newblock \emph{McGill University}, 2018.

\bibitem[Moerland et~al.(2020)Moerland, Broekens, and
  Jonker]{moerland2020model}
Thomas~M Moerland, Joost Broekens, and Catholijn~M Jonker.
\newblock Model-based reinforcement learning: A survey.
\newblock \emph{arXiv preprint arXiv:2006.16712}, 2020.

\bibitem[Sutton and Barto(2018)]{SuttonBarto-39}
Richard~S. Sutton and Andrew~G. Barto.
\newblock \emph{Reinforcement learning: An introduction}.
\newblock MIT press, 2018.

\bibitem[Vinyals et~al.(2019)Vinyals, Babuschkin, Czarnecki, Mathieu, Dudzik,
  Chung, Choi, Powell, Ewalds, Georgiev, et~al.]{vinyals2019grandmaster}
Oriol Vinyals, Igor Babuschkin, Wojciech~M Czarnecki, Micha{\"e}l Mathieu,
  Andrew Dudzik, Junyoung Chung, David~H Choi, Richard Powell, Timo Ewalds,
  Petko Georgiev, et~al.
\newblock Grandmaster level in starcraft ii using multi-agent reinforcement
  learning.
\newblock \emph{Nature}, 575\penalty0 (7782):\penalty0 350--354, 2019.

\bibitem[Jaderberg et~al.(2019)Jaderberg, Czarnecki, Dunning, Marris, Lever,
  Castaneda, Beattie, Rabinowitz, Morcos, Ruderman, et~al.]{jaderberg2019human}
Max Jaderberg, Wojciech~M Czarnecki, Iain Dunning, Luke Marris, Guy Lever,
  Antonio~Garcia Castaneda, Charles Beattie, Neil~C Rabinowitz, Ari~S Morcos,
  Avraham Ruderman, et~al.
\newblock Human-level performance in 3d multiplayer games with population-based
  reinforcement learning.
\newblock \emph{Science}, 364\penalty0 (6443):\penalty0 859--865, 2019.

\bibitem[Mnih et~al.(2016)Mnih, Badia, Mirza, Graves, Lillicrap, Harley,
  Silver, and Kavukcuoglu]{MnihBadia-14}
Volodymyr Mnih, Adria~Puigdomenech Badia, Mehdi Mirza, Alex Graves, Timothy
  Lillicrap, Tim Harley, David Silver, and Koray Kavukcuoglu.
\newblock Asynchronous methods for deep reinforcement learning.
\newblock In \emph{International conference on machine learning}, International
  conference on machine learning, pages 1928--1937. PMLR, 2016.

\bibitem[Schulman et~al.(2015)Schulman, Levine, Abbeel, Jordan, and
  Moritz]{SchulmanLevine-34}
John Schulman, Sergey Levine, Pieter Abbeel, Michael Jordan, and Philipp
  Moritz.
\newblock Trust region policy optimization.
\newblock In \emph{International conference on machine learning}, International
  conference on machine learning, pages 1889--1897. PMLR, 2015.

\bibitem[Schulman et~al.(2017)Schulman, Wolski, Dhariwal, Radford, and
  Klimov]{SchulmanWolski-15}
John Schulman, Filip Wolski, Prafulla Dhariwal, Alec Radford, and Oleg Klimov.
\newblock Proximal policy optimization algorithms.
\newblock \emph{arXiv preprint arXiv:1707.06347}, 2017.

\bibitem[Mnih et~al.(2013)Mnih, Kavukcuoglu, Silver, Graves, Antonoglou,
  Wierstra, and Riedmiller]{MnihKavukcuoglu-16}
Volodymyr Mnih, Koray Kavukcuoglu, David Silver, Alex Graves, Ioannis
  Antonoglou, Daan Wierstra, and Martin Riedmiller.
\newblock Playing atari with deep reinforcement learning.
\newblock \emph{arXiv preprint arXiv:1312.5602}, 2013.

\bibitem[Mnih et~al.(2015)Mnih, Kavukcuoglu, Silver, Rusu, Veness, Bellemare,
  Graves, Riedmiller, Fidjeland, and Ostrovski]{MnihKavukcuoglu-17}
Volodymyr Mnih, Koray Kavukcuoglu, David Silver, Andrei~A. Rusu, Joel Veness,
  Marc~G. Bellemare, Alex Graves, Martin Riedmiller, Andreas~K. Fidjeland, and
  Georg Ostrovski.
\newblock Human-level control through deep reinforcement learning.
\newblock \emph{nature}, 518\penalty0 (7540):\penalty0 529--533, 2015.

\bibitem[Bellemare et~al.(2017)Bellemare, Dabney, and
  Munos]{BellemareDabney-37}
Marc~G. Bellemare, Will Dabney, and Rémi Munos.
\newblock A distributional perspective on reinforcement learning.
\newblock In \emph{International Conference on Machine Learning}, International
  Conference on Machine Learning, pages 449--458. PMLR, 2017.

\bibitem[Dabney et~al.(2018)Dabney, Rowland, Bellemare, and
  Munos]{DabneyRowland-38}
Will Dabney, Mark Rowland, Marc~G. Bellemare, and Rémi Munos.
\newblock Distributional reinforcement learning with quantile regression.
\newblock In \emph{Thirty-Second AAAI Conference on Artificial Intelligence},
  Thirty-Second AAAI Conference on Artificial Intelligence, 2018.

\bibitem[Andrychowicz et~al.(2017)Andrychowicz, Wolski, Ray, Schneider, Fong,
  Welinder, McGrew, Tobin, Abbeel, and Zaremba]{AndrychowiczWolski-18}
Marcin Andrychowicz, Filip Wolski, Alex Ray, Jonas Schneider, Rachel Fong,
  Peter Welinder, Bob McGrew, Josh Tobin, Pieter Abbeel, and Wojciech Zaremba.
\newblock Hindsight experience replay.
\newblock \emph{arXiv preprint arXiv:1707.01495}, 2017.

\bibitem[Lillicrap et~al.(2015)Lillicrap, Hunt, Pritzel, Heess, Erez, Tassa,
  Silver, and Wierstra]{LillicrapHunt-35}
Timothy~P. Lillicrap, Jonathan~J. Hunt, Alexander Pritzel, Nicolas Heess, Tom
  Erez, Yuval Tassa, David Silver, and Daan Wierstra.
\newblock Continuous control with deep reinforcement learning.
\newblock \emph{arXiv preprint arXiv:1509.02971}, 2015.

\bibitem[Fujimoto et~al.(2018)Fujimoto, Hoof, and Meger]{FujimotoHoof-36}
Scott Fujimoto, Herke Hoof, and David Meger.
\newblock Addressing function approximation error in actor-critic methods.
\newblock In \emph{International Conference on Machine Learning}, International
  Conference on Machine Learning, pages 1587--1596. PMLR, 2018.

\bibitem[Haarnoja et~al.(2018)Haarnoja, Zhou, Abbeel, and
  Levine]{HaarnojaZhou-19}
Tuomas Haarnoja, Aurick Zhou, Pieter Abbeel, and Sergey Levine.
\newblock Soft actor-critic: Off-policy maximum entropy deep reinforcement
  learning with a stochastic actor.
\newblock In \emph{International conference on machine learning}, International
  conference on machine learning, pages 1861--1870. PMLR, 2018.

\bibitem[Fortunato et~al.(2017)Fortunato, Azar, Piot, Menick, Osband, Graves,
  Mnih, Munos, Hassabis, and Pietquin]{FortunatoAzar-47}
Meire Fortunato, Mohammad~Gheshlaghi Azar, Bilal Piot, Jacob Menick, Ian
  Osband, Alex Graves, Vlad Mnih, Remi Munos, Demis Hassabis, and Olivier
  Pietquin.
\newblock Noisy networks for exploration.
\newblock \emph{arXiv preprint arXiv:1706.10295}, 2017.

\bibitem[Bellemare et~al.(2016)Bellemare, Srinivasan, Ostrovski, Schaul,
  Saxton, and Munos]{BellemareSrinivasan-40}
Marc Bellemare, Sriram Srinivasan, Georg Ostrovski, Tom Schaul, David Saxton,
  and Remi Munos.
\newblock Unifying count-based exploration and intrinsic motivation.
\newblock \emph{Advances in neural information processing systems},
  29:\penalty0 1471--1479, 2016.

\bibitem[Ostrovski et~al.(2017)Ostrovski, Bellemare, Oord, and
  Munos]{OstrovskiBellemare-41}
Georg Ostrovski, Marc~G. Bellemare, Aäron Oord, and Rémi Munos.
\newblock Count-based exploration with neural density models.
\newblock In \emph{International conference on machine learning}, International
  conference on machine learning, pages 2721--2730. PMLR, 2017.

\bibitem[Tang et~al.(2017)Tang, Houthooft, Foote, Stooke, Chen, Duan, Schulman,
  De~Turck, and Abbeel]{TangHouthooft-42}
Haoran Tang, Rein Houthooft, Davis Foote, Adam Stooke, Xi~Chen, Yan Duan, John
  Schulman, Filip De~Turck, and Pieter Abbeel.
\newblock \# exploration: A study of count-based exploration for deep
  reinforcement learning.
\newblock In \emph{31st Conference on Neural Information Processing Systems
  (NIPS)}, volume~30 of \emph{31st Conference on Neural Information Processing
  Systems (NIPS)}, pages 1--18, 2017.

\bibitem[Oudeyer et~al.(2007)Oudeyer, Kaplan, and Hafner]{OudeyerKaplan-43}
Pierre-Yves Oudeyer, Frdric Kaplan, and Verena~V. Hafner.
\newblock Intrinsic motivation systems for autonomous mental development.
\newblock \emph{IEEE transactions on evolutionary computation}, 11\penalty0
  (2):\penalty0 265--286, 2007.

\bibitem[Pathak et~al.(2017)Pathak, Agrawal, Efros, and
  Darrell]{PathakAgrawal-21}
Deepak Pathak, Pulkit Agrawal, Alexei~A. Efros, and Trevor Darrell.
\newblock Curiosity-driven exploration by self-supervised prediction.
\newblock In \emph{International conference on machine learning}, International
  conference on machine learning, pages 2778--2787. PMLR, 2017.

\bibitem[Burda et~al.(2018)Burda, Edwards, Storkey, and
  Klimov]{BurdaEdwards-22}
Yuri Burda, Harrison Edwards, Amos Storkey, and Oleg Klimov.
\newblock Exploration by random network distillation.
\newblock \emph{arXiv preprint arXiv:1810.12894}, 2018.

\bibitem[Pathak et~al.(2019)Pathak, Gandhi, and Gupta]{PathakGandhi-49}
Deepak Pathak, Dhiraj Gandhi, and Abhinav Gupta.
\newblock Self-supervised exploration via disagreement.
\newblock In \emph{International conference on machine learning}, International
  conference on machine learning, pages 5062--5071. PMLR, 2019.

\bibitem[Badia et~al.(2020{\natexlab{a}})Badia, Sprechmann, Vitvitskyi, Guo,
  Piot, Kapturowski, Tieleman, Arjovsky, Pritzel, and Bolt]{BadiaSprechmann-50}
Adrià~Puigdomènech Badia, Pablo Sprechmann, Alex Vitvitskyi, Daniel Guo,
  Bilal Piot, Steven Kapturowski, Olivier Tieleman, Martín Arjovsky, Alexander
  Pritzel, and Andew Bolt.
\newblock Never give up: Learning directed exploration strategies.
\newblock \emph{arXiv preprint arXiv:2002.06038}, 2020{\natexlab{a}}.

\bibitem[Badia et~al.(2020{\natexlab{b}})Badia, Piot, Kapturowski, Sprechmann,
  Vitvitskyi, Guo, and Blundell]{BadiaPiot-2}
Adrià~Puigdomènech Badia, Bilal Piot, Steven Kapturowski, Pablo Sprechmann,
  Alex Vitvitskyi, Zhaohan~Daniel Guo, and Charles Blundell.
\newblock Agent57: Outperforming the atari human benchmark.
\newblock In \emph{International Conference on Machine Learning}, International
  Conference on Machine Learning, pages 507--517. PMLR, 2020{\natexlab{b}}.

\bibitem[Ecoffet et~al.(2021)Ecoffet, Huizinga, Lehman, Stanley, and
  Clune]{EcoffetHuizinga-7}
Adrien Ecoffet, Joost Huizinga, Joel Lehman, Kenneth~O. Stanley, and Jeff
  Clune.
\newblock First return, then explore.
\newblock \emph{Nature}, 590\penalty0 (7847):\penalty0 580--586, 2021.

\bibitem[Maurer et~al.(2016)Maurer, Pontil, and
  Romera-Paredes]{MaurerPontil-53}
Andreas Maurer, Massimiliano Pontil, and Bernardino Romera-Paredes.
\newblock The benefit of multitask representation learning.
\newblock \emph{Journal of Machine Learning Research}, 17\penalty0
  (81):\penalty0 1--32, 2016.

\bibitem[Vithayathil~Varghese and Mahmoud(2020)]{VithayathilVargheseMahmoud-54}
Nelson Vithayathil~Varghese and Qusay~H. Mahmoud.
\newblock A survey of multi-task deep reinforcement learning.
\newblock \emph{Electronics}, 9\penalty0 (9):\penalty0 1363, 2020.

\bibitem[Rusu et~al.(2016)Rusu, Rabinowitz, Desjardins, Soyer, Kirkpatrick,
  Kavukcuoglu, Pascanu, and Hadsell]{RusuRabinowitz-56}
Andrei~A. Rusu, Neil~C. Rabinowitz, Guillaume Desjardins, Hubert Soyer, James
  Kirkpatrick, Koray Kavukcuoglu, Razvan Pascanu, and Raia Hadsell.
\newblock Progressive neural networks.
\newblock \emph{arXiv preprint arXiv:1606.04671}, 2016.

\bibitem[Fernando et~al.(2017)Fernando, Banarse, Blundell, Zwols, Ha, Rusu,
  Pritzel, and Wierstra]{FernandoBanarse-55}
Chrisantha Fernando, Dylan Banarse, Charles Blundell, Yori Zwols, David Ha,
  Andrei~A. Rusu, Alexander Pritzel, and Daan Wierstra.
\newblock Pathnet: Evolution channels gradient descent in super neural
  networks.
\newblock \emph{arXiv preprint arXiv:1701.08734}, 2017.

\bibitem[Rusu et~al.(2015)Rusu, Colmenarejo, Gulcehre, Desjardins, Kirkpatrick,
  Pascanu, Mnih, Kavukcuoglu, and Hadsell]{RusuColmenarejo-57}
Andrei~A. Rusu, Sergio~Gomez Colmenarejo, Caglar Gulcehre, Guillaume
  Desjardins, James Kirkpatrick, Razvan Pascanu, Volodymyr Mnih, Koray
  Kavukcuoglu, and Raia Hadsell.
\newblock Policy distillation.
\newblock \emph{arXiv preprint arXiv:1511.06295}, 2015.

\bibitem[Teh et~al.(2017)Teh, Bapst, Czarnecki, Quan, Kirkpatrick, Hadsell,
  Heess, and Pascanu]{TehBapst-58}
Yee~Whye Teh, Victor Bapst, Wojciech~Marian Czarnecki, John Quan, James
  Kirkpatrick, Raia Hadsell, Nicolas Heess, and Razvan Pascanu.
\newblock Distral: Robust multitask reinforcement learning.
\newblock \emph{arXiv preprint arXiv:1707.04175}, 2017.

\bibitem[Hessel et~al.(2019)Hessel, Soyer, Espeholt, Czarnecki, Schmitt, and
  van Hasselt]{HesselSoyer-59}
Matteo Hessel, Hubert Soyer, Lasse Espeholt, Wojciech Czarnecki, Simon Schmitt,
  and Hado van Hasselt.
\newblock Multi-task deep reinforcement learning with popart.
\newblock In \emph{Proceedings of the AAAI Conference on Artificial
  Intelligence}, volume~33 of \emph{Proceedings of the AAAI Conference on
  Artificial Intelligence}, pages 3796--3803, 2019.
\newblock 01.

\bibitem[Espeholt et~al.(2018)Espeholt, Soyer, Munos, Simonyan, Mnih, Ward,
  Doron, Firoiu, Harley, and Dunning]{EspeholtSoyer-60}
Lasse Espeholt, Hubert Soyer, Remi Munos, Karen Simonyan, Vlad Mnih, Tom Ward,
  Yotam Doron, Vlad Firoiu, Tim Harley, and Iain Dunning.
\newblock Impala: Scalable distributed deep-rl with importance weighted
  actor-learner architectures.
\newblock In \emph{International Conference on Machine Learning}, International
  Conference on Machine Learning, pages 1407--1416. PMLR, 2018.

\bibitem[Florensa et~al.(2017)Florensa, Duan, and Abbeel]{FlorensaDuan-76}
Carlos Florensa, Yan Duan, and Pieter Abbeel.
\newblock Stochastic neural networks for hierarchical reinforcement learning.
\newblock \emph{arXiv preprint arXiv:1704.03012}, 2017.

\bibitem[Frans et~al.(2017)Frans, Ho, Chen, Abbeel, and Schulman]{FransHo-77}
Kevin Frans, Jonathan Ho, Xi~Chen, Pieter Abbeel, and John Schulman.
\newblock Meta learning shared hierarchies.
\newblock \emph{arXiv preprint arXiv:1710.09767}, 2017.

\bibitem[Levy et~al.(2017)Levy, Konidaris, Platt, and Saenko]{LevyKonidaris-78}
Andrew Levy, George Konidaris, Robert Platt, and Kate Saenko.
\newblock Learning multi-level hierarchies with hindsight.
\newblock \emph{arXiv preprint arXiv:1712.00948}, 2017.

\bibitem[Song et~al.(2019)Song, Wang, Lukasiewicz, Xu, and Xu]{SongWang-79}
Yuhang Song, Jianyi Wang, Thomas Lukasiewicz, Zhenghua Xu, and Mai Xu.
\newblock Diversity-driven extensible hierarchical reinforcement learning.
\newblock In \emph{Proceedings of the AAAI conference on artificial
  intelligence}, volume~33 of \emph{Proceedings of the AAAI conference on
  artificial intelligence}, pages 4992--4999, 2019.
\newblock 01.

\bibitem[Sukhbaatar et~al.(2018)Sukhbaatar, Denton, Szlam, and
  Fergus]{SukhbaatarDenton-80}
Sainbayar Sukhbaatar, Emily Denton, Arthur Szlam, and Rob Fergus.
\newblock Learning goal embeddings via self-play for hierarchical reinforcement
  learning.
\newblock \emph{arXiv preprint arXiv:1811.09083}, 2018.

\bibitem[Li et~al.(2019)Li, Wang, Tang, and Zhang]{LiWang-81}
Siyuan Li, Rui Wang, Minxue Tang, and Chongjie Zhang.
\newblock Hierarchical reinforcement learning with advantage-based auxiliary
  rewards.
\newblock \emph{arXiv preprint arXiv:1910.04450}, 2019.

\bibitem[Mishra et~al.(2017)Mishra, Rohaninejad, Chen, and
  Abbeel]{MishraRohaninejad-61}
Nikhil Mishra, Mostafa Rohaninejad, Xi~Chen, and Pieter Abbeel.
\newblock A simple neural attentive meta-learner.
\newblock \emph{arXiv preprint arXiv:1707.03141}, 2017.

\bibitem[Wang et~al.(2016)Wang, Kurth-Nelson, Tirumala, Soyer, Leibo, Munos,
  Blundell, Kumaran, and Botvinick]{WangKurth-Nelson-62}
Jane~X. Wang, Zeb Kurth-Nelson, Dhruva Tirumala, Hubert Soyer, Joel~Z. Leibo,
  Remi Munos, Charles Blundell, Dharshan Kumaran, and Matt Botvinick.
\newblock Learning to reinforcement learn.
\newblock \emph{arXiv preprint arXiv:1611.05763}, 2016.

\bibitem[Finn et~al.(2017)Finn, Abbeel, and Levine]{FinnAbbeel-28}
Chelsea Finn, Pieter Abbeel, and Sergey Levine.
\newblock Model-agnostic meta-learning for fast adaptation of deep networks.
\newblock In \emph{International Conference on Machine Learning}, International
  Conference on Machine Learning, pages 1126--1135. PMLR, 2017.

\bibitem[Nichol et~al.(2018)Nichol, Achiam, and Schulman]{NicholAchiam-64}
Alex Nichol, Joshua Achiam, and John Schulman.
\newblock On first-order meta-learning algorithms.
\newblock \emph{arXiv preprint arXiv:1803.02999}, 2018.

\bibitem[Yoon et~al.(2018)Yoon, Kim, Dia, Kim, Bengio, and Ahn]{YoonKim-65}
Jaesik Yoon, Taesup Kim, Ousmane Dia, Sungwoong Kim, Yoshua Bengio, and Sungjin
  Ahn.
\newblock Bayesian model-agnostic meta-learning.
\newblock In \emph{Proceedings of the 32nd International Conference on Neural
  Information Processing Systems}, Proceedings of the 32nd International
  Conference on Neural Information Processing Systems, pages 7343--7353, 2018.

\bibitem[Rajeswaran et~al.(2019)Rajeswaran, Finn, Kakade, and
  Levine]{RajeswaranFinn-66}
Aravind Rajeswaran, Chelsea Finn, Sham Kakade, and Sergey Levine.
\newblock Meta-learning with implicit gradients.
\newblock 2019.

\bibitem[Sutton(1991)]{Sutton-68}
Richard~S. Sutton.
\newblock Dyna, an integrated architecture for learning, planning, and
  reacting.
\newblock \emph{ACM Sigart Bulletin}, 2\penalty0 (4):\penalty0 160--163, 1991.

\bibitem[Kurutach et~al.(2018)Kurutach, Clavera, Duan, Tamar, and
  Abbeel]{KurutachClavera-70}
Thanard Kurutach, Ignasi Clavera, Yan Duan, Aviv Tamar, and Pieter Abbeel.
\newblock Model-ensemble trust-region policy optimization.
\newblock \emph{arXiv preprint arXiv:1802.10592}, 2018.

\bibitem[Kalweit and Boedecker(2017)]{KalweitBoedecker-71}
Gabriel Kalweit and Joschka Boedecker.
\newblock Uncertainty-driven imagination for continuous deep reinforcement
  learning.
\newblock In \emph{Conference on Robot Learning}, Conference on Robot Learning,
  pages 195--206. PMLR, 2017.

\bibitem[Gu et~al.(2016)Gu, Lillicrap, Sutskever, and Levine]{GuLillicrap-72}
Shixiang Gu, Timothy Lillicrap, Ilya Sutskever, and Sergey Levine.
\newblock Continuous deep q-learning with model-based acceleration.
\newblock In \emph{International conference on machine learning}, International
  conference on machine learning, pages 2829--2838. PMLR, 2016.

\bibitem[Racanière et~al.(2017)Racanière, Weber, Reichert, Buesing, Guez,
  Rezende, Badia, Vinyals, Heess, and Li]{RacaniereWeber-74}
Sébastien Racanière, Théophane Weber, David~P. Reichert, Lars Buesing,
  Arthur Guez, Danilo Rezende, Adria~Puigdomenech Badia, Oriol Vinyals, Nicolas
  Heess, and Yujia Li.
\newblock Imagination-augmented agents for deep reinforcement learning.
\newblock In \emph{Proceedings of the 31st International Conference on Neural
  Information Processing Systems}, Proceedings of the 31st International
  Conference on Neural Information Processing Systems, pages 5694--5705, 2017.

\bibitem[Feinberg et~al.(2018)Feinberg, Wan, Stoica, Jordan, Gonzalez, and
  Levine]{FeinbergWan-73}
Vladimir Feinberg, Alvin Wan, Ion Stoica, Michael~I. Jordan, Joseph~E.
  Gonzalez, and Sergey Levine.
\newblock Model-based value estimation for efficient model-free reinforcement
  learning.
\newblock \emph{arXiv preprint arXiv:1803.00101}, 2018.

\bibitem[Buckman et~al.(2018)Buckman, Hafner, Tucker, Brevdo, and
  Lee]{BuckmanHafner-75}
Jacob Buckman, Danijar Hafner, George Tucker, Eugene Brevdo, and Honglak Lee.
\newblock Sample-efficient reinforcement learning with stochastic ensemble
  value expansion.
\newblock \emph{arXiv preprint arXiv:1807.01675}, 2018.

\bibitem[Furuta et~al.(2021)Furuta, Matsushima, Kozuno, Matsuo, Levine, Nachum,
  and Gu]{furuta2021policy}
Hiroki Furuta, Tatsuya Matsushima, Tadashi Kozuno, Yutaka Matsuo, Sergey
  Levine, Ofir Nachum, and Shixiang~Shane Gu.
\newblock Policy information capacity: Information-theoretic measure for task
  complexity in deep reinforcement learning.
\newblock \emph{arXiv preprint arXiv:2103.12726}, 2021.

\end{thebibliography}






\end{document}